\documentclass[11pt]{article}
\usepackage{acl}  
\usepackage{times}
\usepackage{latexsym}
\usepackage[T1]{fontenc}
\usepackage[utf8]{inputenc}
\usepackage{microtype}
\usepackage{inconsolata}
\usepackage{amsmath,amssymb,amsthm}
\usepackage{mathtools}
\usepackage{bm}
\usepackage{graphicx}
\usepackage{booktabs}
\usepackage{multirow}
\usepackage{xcolor}
\usepackage{cleveref}
\usepackage{enumitem}
\usepackage{subcaption}
\usepackage{float}
\usepackage{xspace}

\newcommand{\R}{\mathbb{R}}

\newcommand{\Tr}{\mathrm{Tr}}

\newcommand{\cov}{\mathrm{Cov}}
\newcommand{\wca}{\textsc{wca}\xspace}
\newcommand{\scm}{\textsc{scm}\xspace}
\newcommand{\dom}{difference-of-means\xspace}

\title{Concepts Whisper:\\Spectral Anti-Concentration and the Dual Geometry\\of Transformer Representations}
\author{Pratyush Acharya$^{1*}$ \quad Nuraj Rimal$^{2*}$ \quad Habish Dhakal$^{1*}$ \\
  $^{1}$SecurityPal AI \qquad $^{2}$Southern Illinois University Edwardsville \\
  \texttt{\{pratyush.acharya,habish.dhakal\}@securitypalhq.com} \\
  \texttt{nurimal@siue.edu}}
\date{}

\begin{document}
\maketitle

\begingroup
\renewcommand{\thefootnote}{\fnsymbol{footnote}}
\footnotetext[1]{Equal contribution.}
\endgroup

\begin{abstract}
We find that transformer concept representations systematically \emph{anti-concentrate} in the spectral tail of the unembedding covariance, encoding word-level concepts in low-variance directions across a 17-model core suite and an expanded set of 22 semantic concept categories, with convergent replications from three independent extraction methods. Residual-stream difference-of-means vectors anti-concentrate in all 17 models (model-level one-sample $t$-test, $p = 3.8\times 10^{-9}$), and remain more tail-aligned than norm-matched random directions in 13 of 17; convergent support comes from sparse autoencoder (SAE) features ($p = 4.5\times 10^{-19}$ across concepts within a model) and linear probes on Llama and Qwen. We identify a \emph{dual geometry}: activation-space concept directions anti-concentrate while static unembedding-row contrasts \emph{concentrate} in high-variance directions ($p < 10^{-4}$). This investigation arose from testing whether the causal inner product of \citet{park2024linear} aids cross-lingual concept transport; a matched-spectrum randomization across 17 models and four language pairs finds no evidence that Whitened Causal Alignment improves over spectral regularization alone ($p = 0.95$). Split-injection interventions, restricted to steering strengths at which both arms remain interpretable, show the predicted interference asymmetry in four of five models (paired Cohen's $d_z$ up to $1.19$) with no significant reversal inside that regime, and POS-tag probing across eight models shows syntax preferentially encoded in the high-variance subspace in six of eight architectures on English, with a significant reversal in the Qwen~2.5 family. These results suggest transformers rotate semantic content into spectrally quiet regions during contextualized processing, where, in some architectures, interventions may reduce grammatical disruption relative to high-variance steering.
\end{abstract}

\section{Introduction}
\label{sec:intro}

\begin{figure*}[t]
  \centering
  \includegraphics[width=\textwidth]{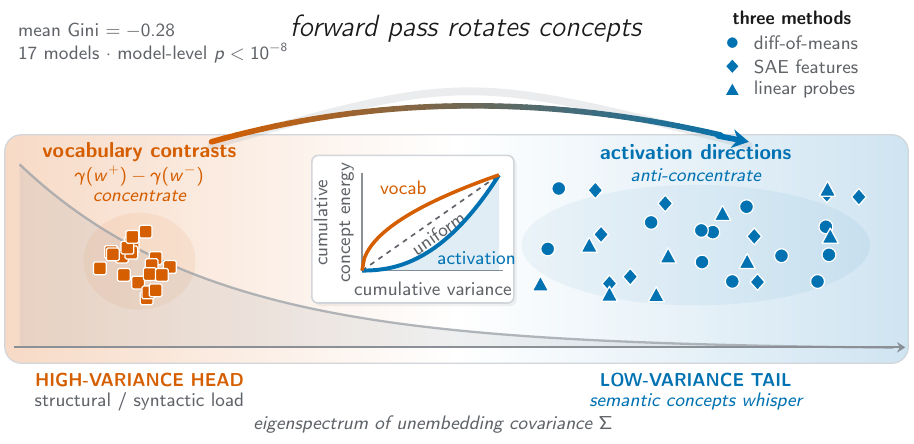}
  \caption{\textbf{The dual geometry of transformer representations.}
  We order the eigenvectors of the unembedding covariance $\Sigma$ from the
  high-variance \emph{head} (left) to the low-variance \emph{tail} (right).
  Concept directions built from the \emph{static vocabulary}---unembedding-row
  contrasts $\gamma(w^{+})-\gamma(w^{-})$ (orange)---\emph{concentrate} in the
  high-variance head, whereas \emph{contextualized activation}-space concept
  directions recovered by three independent methods (difference-of-means, SAE
  features, and linear probes; blue) \emph{anti-concentrate} in the low-variance
  tail. During the forward pass, semantic concept information is thus rotated
  out of the loud head and into the quiet tail (mean Gini deviation $=-0.28$
  across 17 models, model-level $p<10^{-8}$). Inset: the cumulative concept-energy curve
  bows \emph{below} the uniform diagonal for activation directions
  (anti-concentration) and above it for vocabulary contrasts (concentration).
  The high-variance head disproportionately carries structural (syntactic)
  load---a functional reading that holds on most, though not all, tested models
  (\S\ref{sec:why})---while word-level semantic concepts occupy the low-variance
  tail.}
  \label{fig:dual-geometry-intro}
\end{figure*}

The mechanistic interpretability of large language models (LLMs) rests on an empirical observation: high-level concepts appear to be encoded as linear directions in activation space~\citep{mikolov2013efficient,marks2024geometry,tigges2024linear,nanda2023emergent}. \citet{park2024linear} formalize this \emph{Linear Representation Hypothesis} (LRH) and argue the natural geometry is a \emph{causal inner product}---a whitening of the representation space by the unembedding covariance---under which causally separable concepts are provably orthogonal~\citep{park2024linear,park2025geometry}. In parallel, the Platonic Representation Hypothesis~\citep{huh2024platonic} and universal-feature work~\citep{lan2024sparse} suggest that capable models converge on shared representational structure, motivating \emph{cross-lingual concept transport} via Procrustes alignment (the optimal orthogonal rotation matching two sets of points)~\citep{conneau2018word,conneau2020emerging}.

The intersection of these hypotheses motivates what we call \emph{Whitened Causal Alignment} (\wca): align two representation spaces after whitening with the inverse square root of each model's unembedding covariance. If the causal geometry hypothesis is correct, \wca should outperform naive Procrustes by stripping language-specific anisotropy (the tendency of activations to crowd into a narrow cone)~\citep{lim2025language,wendler2024llamas}.

We set out to validate this. Across 17 models and 4 language pairs, the causal eigendirections do \emph{not} measurably improve transport in a matched-spectrum test (\Cref{sec:wca}). But this failure exposed something more interesting: activation-space concept vectors systematically \emph{anti-concentrate} in the eigenspectrum of the unembedding covariance $\Sigma$ (\Cref{eq:cov}), preferentially encoding in low-variance directions (\Cref{sec:anti-concentration}). The pattern is not an artifact: it persists for SAE features, linear probes, and \dom vectors (\Cref{sec:artifact-check}). Moreover, the \emph{static vocabulary} shows the opposite pattern---concept-specific unembedding-row differences \emph{concentrate} in high-variance directions, exposing a \emph{dual geometry} (\Cref{fig:dual-geometry-intro}; \Cref{sec:dual-geometry}). Steering experiments and POS-tag probing provide functional support for why this dual geometry exists (\Cref{sec:why}). Beyond interpretability, the finding bears on NLP practice: PCA- and low-rank-based analyses can miss concept structure that lives in the spectral tail; steering vectors projected into that tail may edit meaning with less collateral damage to fluency; and it offers a spectral account of capacity allocation---why semantic features are driven into superposition even at large $d$ (\Cref{sec:discussion}).

\paragraph{Contributions.}
(i)~A negative result for causal eigendirections in cross-lingual transport, established by matched-spectrum randomization across 17 models. (ii)~Discovery and validation of spectral anti-concentration of activation-space concept representations across three independent extraction methods. (iii)~Identification of a dual geometry between vocabulary-space concentration and activation-space anti-concentration, with a random-pair null model showing concept-specific structure. (iv)~Functional evidence via split-injection steering across five models and POS-tag probing across eight models; \Cref{tab:claims} maps each claim to its evidence and strength. Code and data to reproduce every experiment---the full extraction and analysis pipeline, all concept-pair files, and exact hyperparameters and seeds---are publicly available at \url{https://github.com/pratyush-acharya/spectral_anti_concentration}.

\begin{table}[t]
\centering
\small
\caption{Map of central claims to their evidence and the strength we ascribe to each. We separate \emph{robust} findings (consistent within the model suite tested for that claim, with strong statistics) from \emph{partial} ones (architecture-dependent or based on a smaller subset).}
\label{tab:claims}
\resizebox{\columnwidth}{!}{%
\begin{tabular}{p{0.36\columnwidth}p{0.36\columnwidth}c}
\toprule
\textbf{Claim} & \textbf{Evidence} & \textbf{Strength} \\
\midrule
Anti-concentration of activation-space concepts & 17/17 models negative vs.\ the uniform baseline (model-level $p=3.8\times10^{-9}$); 13/17 vs.\ the matched random-direction null; SAE/probe replications (\S\ref{sec:anti-concentration},\ref{sec:artifact-check}) & robust (13/17 vs.\ matched null) \\
Independent of background spectrum & $r=0.019$ between random and concept \scm (\S\ref{sec:anti-concentration}) & suggestive ($n{=}17$) \\
Dual geometry (vocabulary concentrates) & 4 models, random-pair null $p<10^{-4}$ (\S\ref{sec:dual-geometry}) & robust \\
WCA adds no eigendirection benefit & 17 models $\times$ 4 language pairs, $p=0.95$ (\S\ref{sec:wca}) & robust (null) \\
Interference asymmetry (steering) & 4/5 models show a significant in-window shout$>$whisper asymmetry (7 of 13 configurations, nested within those 4 models; sweep $p$ uncorrected); Mistral undetermined. Magnitude is $\alpha$-dependent (\S\ref{sec:why}) & partial \\
Syntax in high-variance subspace & 6/8 models on English; Qwen~2.5 reverses, as do both models tested on Chinese (\S\ref{sec:why}) & partial \\
\bottomrule
\end{tabular}
}
\end{table}

\section{Background and Related Work}
\label{sec:background}

\paragraph{Linear representations and the causal inner product.}
Linear encoding has been demonstrated for truth~\citep{marks2024geometry,burns2023discovering}, sentiment~\citep{tigges2024linear}, world state~\citep{nanda2023emergent}, and refusal~\citep{arditi2024refusal}; \citet{engels2025not} note that some features are multi-dimensional. For a token set $\mathcal{L}$, we form the regularized \emph{uncentered} second moment of the unembedding rows,
\begin{equation}
  \Sigma_{\mathcal{L}} = \tfrac{1}{|\mathcal{L}|}\sum_{w\in\mathcal{L}} \gamma(w)\gamma(w)^\top + \tau I,
  \label{eq:cov}
\end{equation}
where $\gamma(w)\!\in\!\R^d$ is the unembedding row for $w$ and $\tau$ is a ridge term; throughout, we reserve $\lambda_i$ for the eigenvalues of $\Sigma_\mathcal{L}$. For brevity we refer to $\Sigma_\mathcal{L}$ as the unembedding \emph{covariance} throughout, though as defined it is an uncentered second moment, and we drop the subscript and write $\Sigma$ wherever the token set $\mathcal{L}$ is clear from context. The causal inner product is $\langle u,v\rangle_C=u^\top\Sigma_\mathcal{L}^{-1}v$~\citep{park2024linear,park2025geometry}. \citet{park2024linear} define this via the \emph{centered} covariance $\cov(\gamma)$; we use the uncentered second moment of \Cref{eq:cov} and confirm in \Cref{app:robustness-controls} that centering leaves the effect unchanged (if anything slightly stronger).

\paragraph{Cross-lingual alignment and \wca.}
Procrustes alignment finds $Q$ minimizing $\|X_\text{src}-X_\text{tgt}Q\|_F$ with $Q^\top Q=I$~\citep{conneau2018word}. \wca composes this with whitening: $v_\text{tgt} = \Sigma_\text{tgt}^{1/2}\,Q^\top\,\Sigma_\text{src}^{-1/2}\,v_\text{src}$.

\paragraph{Anisotropy and SAEs.}
Transformer activations are anisotropic, with a few outlier dimensions dominating variance~\citep{ethayarajh2019contextual,gao2019representation,timkey2021all,puccetti2022outlier,kovaleva2021bert,machina2024anisotropy}; some of these dimensions encode structured content~\citep{gurnee2023language}. SAEs~\citep{bricken2023towards,cunningham2024sparse,templeton2024scaling,gao2025scaling,lieberum2024gemma,he2024llama} decompose activations into sparse interpretable features.

\paragraph{Spectra, steering, and whitening.}
\citet{martin2021implicit,martin2019traditional,martin2021predicting} study spectral structure of weights; \citet{staats2024small} show small singular values carry information; \citet{sharma2023truth} show pruning top components can \emph{improve} reasoning. Activation steering~\citep{turner2024activation,panickssery2024steering,zou2023representation,li2024inference,subramani2022extracting} manipulates concept directions at inference time. Whitening helps sentence embeddings~\citep{su2021whitening,huang2021whiteningbert}, plausibly by amplifying low-variance directions where concepts live. \citet{papyan2020prevalence,wu2024linguistic} study neural collapse; our finding is distinct---concepts are pushed \emph{away from} the dominant spectral subspace.

\section{Experimental Setup}
\label{sec:setup}

\paragraph{Models.}
Core spectral experiments use 17 decoder-only base models across multiple families (\Cref{app:models}): Llama~3.2 (1B, 3B); Qwen~2.5 (0.5B--14B); Qwen~3 (0.6B--8B); Gemma~2 (2B, 9B); Mistral-7B-v0.3; SmolLM2-1.7B; and two MoE models (OLMoE-1B-7B, JetMoE-8B). We additionally use Llama~3.1-8B for SAE and functional tests. Hidden dims 896--5120; vocab 32K--256K. All are decoder-only base models.

\paragraph{Concepts.}
27 concept directions in four categories from the counterfactual word-pair dataset of \citet{park2024linear} (sourced from \citealp{gladkova2016analogy}): Verb Morphology (10), Semantic (7), Grammatical (6), and Language Pairs (4); 4--228 pairs each. These are \emph{word-level, predominantly single-token} counterfactual pairs; our claims concern this population, and abstract, relational, or discourse-level concepts remain out of scope (see Limitations). To probe whether the small semantic subset limits generality, we further introduce an \emph{expanded} semantic set of 15 additional BATS categories (encyclopedic and lexicographic), tripling the semantic concepts from 7 to 22 (\Cref{sec:anti-concentration}, \Cref{app:expanded-semantic}).

\paragraph{Three independent extraction methods.}
(1)~\emph{\dom on residuals (subtractive):} $v_\text{dom} = \tfrac{1}{N}\sum_i(h(w_i^+)-h(w_i^-))$ over final-layer pooled activations. As a subtractive estimator it could in principle cancel shared high-variance components, motivating the artifact controls in \Cref{sec:artifact-check}. (2)~\emph{SAE features (contextualized, unsupervised):} top-$k$ differentially activating features from Gemma Scope and Llama Scope; decoder columns serve as concept directions. No subtraction. (3)~\emph{Linear probes (contextualized, supervised):} L2-regularized logistic regression on uncentered final-layer activations; the weight vector is the concept direction. L2 biases toward high-variance directions, making this a conservative test.

\paragraph{Spectral analysis.}
Project concept vectors onto the eigenbasis of $\Sigma_\mathcal{L}$ (\Cref{eq:cov}) with eigenvalues $\lambda_1\!\ge\!\cdots\!\ge\!\lambda_d$. The fractional energy at $u_i$ is $E_i=(v^\top u_i)^2/\|v\|^2$; cumulative energy $C(k)=\sum_{i\le k}E_i$ is plotted against cumulative variance $V(k)=\sum_{i\le k}\lambda_i/\Tr(\Sigma)$ to compare across $d$. \emph{Gini deviation} is the signed area between $C(\cdot)$ and the uniform baseline (negative = anti-concentration). \emph{Spectral Center of Mass} (\scm) is the $V(k)$ at which $C(k)=0.5$; \scm near 1 means encoding in low-eigenvalue directions. For partitioned subspace tests we use $k=\lfloor 0.1d\rfloor$ and verify robustness at $k\in\{5\%,10\%,20\%\}$ (\Cref{app:pos-sensitivity}). Hyperparameters and seeds are listed in \Cref{app:hyperparams}.

\section{The Causal Geometry Does Not Aid Cross-Lingual Transport}
\label{sec:wca}

To determine whether \wca's benefit stems from the specific eigenvector directions of $\Sigma$ or merely from its spectral shape, we construct matched-spectrum \emph{fake} covariances $\Sigma_\text{fake}=V\Lambda V^\top$, where $\Lambda$ contains the true eigenvalues and $V$ is Haar-random orthogonal. We run the full \wca pipeline with $\Sigma_\text{fake}$ for both source and target, with 5 seeds across all 17 models $\times$ 4 language pairs (68 combinations) at $\tau=0.1$.

\begin{figure*}[t]
  \centering
  \includegraphics[width=0.92\textwidth]{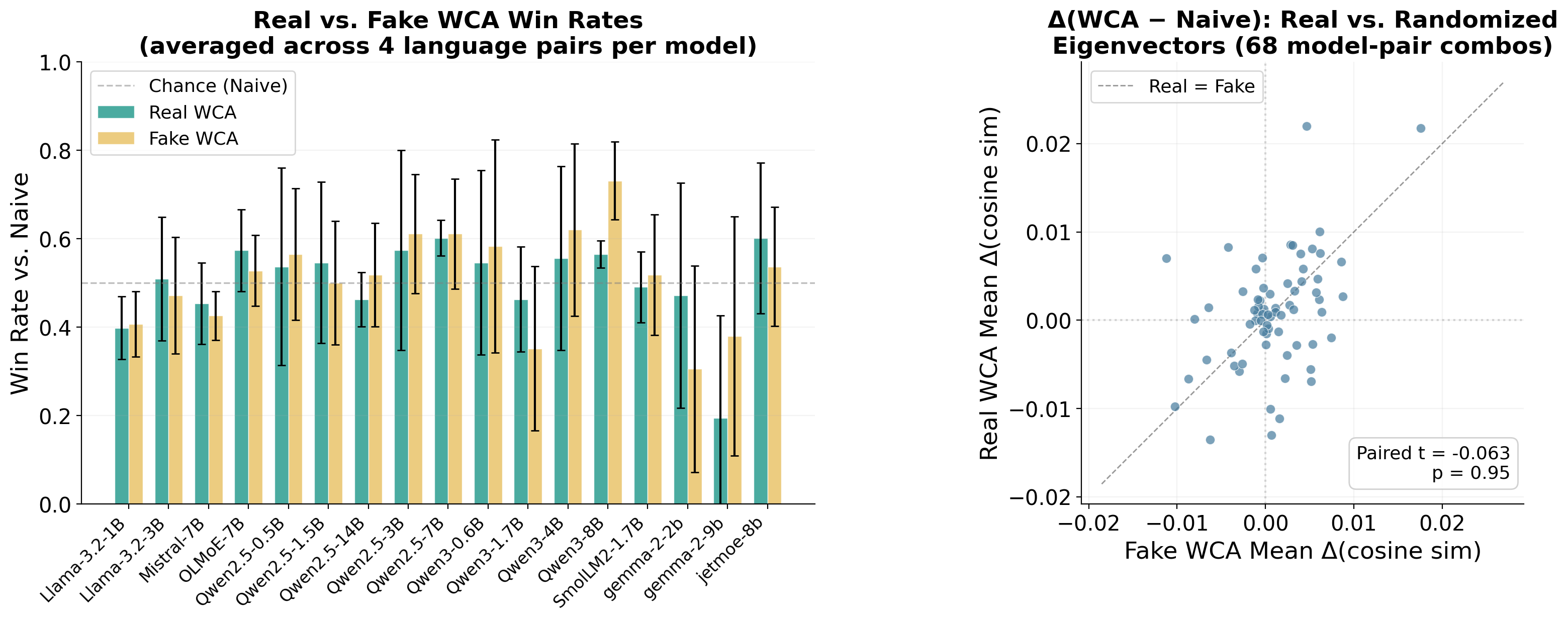}
  \caption{Matched-spectrum randomization across 68 model-pair combinations. Real \wca (50.3\% win rate over naive Procrustes) is indistinguishable from Fake \wca with randomized eigenvectors (51.0\%); paired $t$-test $t=-0.063$, $p=0.95$.}
  \label{fig:randomization-main}
\end{figure*}

Real \wca achieves a $50.3\%$ win rate over naive Procrustes (mean $\Delta=+0.0009$); Fake \wca achieves $51.0\%$ ($\Delta=+0.0010$); paired $t$-test $t=-0.063$, $p=0.95$ (\Cref{fig:randomization-main}). Three of the four language pairs are non-English (es-de, fr-de, fr-es); restricting the test to those, the null still holds (Real \wca $49.2\%$ vs.\ Fake $50.9\%$; real$-$fake $t=-0.49$), so it is not an artifact of always pairing with English. Within this resolution any observed benefit of \wca is attributable to spectral regularization---dampening high-variance directions regardless of semantic content---rather than to semantically meaningful causal eigendirections. This negative result is the direct entry point to our positive findings: if the eigenvectors of $\Sigma$ do not carry cross-lingual transport signal, what \emph{do} they encode?

\section{Spectral Anti-Concentration}
\label{sec:anti-concentration}

\begin{figure}[t]
  \centering
  \includegraphics[width=\columnwidth]{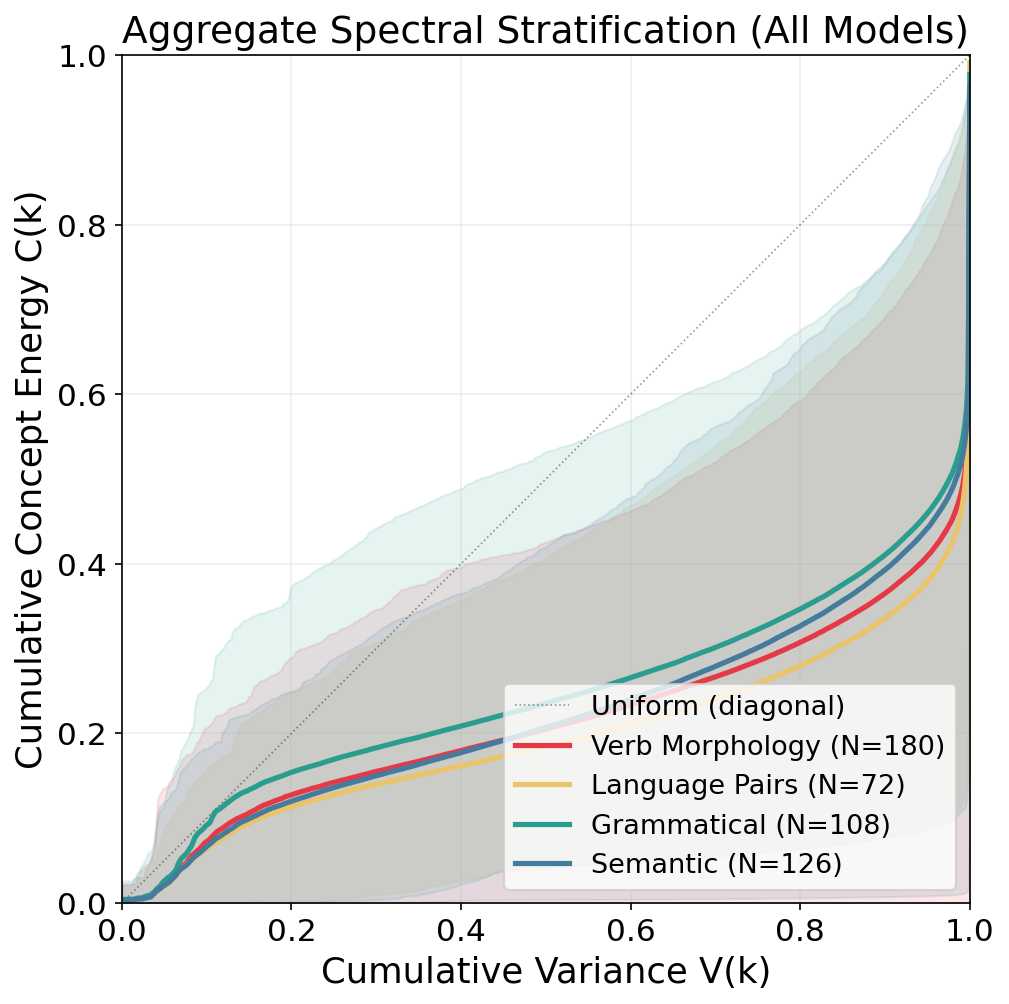}
  \caption{Aggregate spectral energy CDFs across 18 models (the 17-model core suite plus Llama-3.1-8B, hence group $N=27\times18$). Curves fall below the uniform diagonal, indicating anti-concentration; shaded regions show spread across models and concepts, not a random-direction baseline (see \Cref{app:robustness-controls}). $x$-axis: cumulative variance $V(k)$.}
  \label{fig:aggregate-cdf}
\end{figure}

The matched-spectrum pipeline required eigendecompositions of $\Sigma$ for every model, naturally surfacing how concept vectors distribute across these eigenbases. Across all 17 models, \dom concept vectors anti-concentrate (\Cref{fig:aggregate-cdf}). Taking the \emph{model} as the unit of analysis---the level at which our observations are independent---the mean Gini deviation is $-0.282$ and all 17 models are negative (one-sample $t=-11.5$, $p=3.8\times10^{-9}$, $n=17$). Every concept category is significant on the same basis and negative in all 17 models, the weakest being Grammatical ($t=-9.5$, $p=5.7\times10^{-8}$). Within-model tests over the 27 concepts are far stronger (all $p<10^{-6}$), but concepts in a model share an eigenbasis and are not independent replicates, so the model-level test is our headline and concept-level pooling is descriptive only. The \scm gap between concept and random baselines is $+0.169$ on average (\Cref{tab:scm-summary}), and is independent of background spectral shape (random vs.\ concept \scm: $r=0.019$, $p=0.94$). Pairwise Wilcoxon tests find no significant differences between most concept categories (e.g., Verb Morphology vs.\ Semantic, $p=0.73$); all categories anti-concentrate to comparable degree.

These Gini deviations are measured against the \emph{uniform diagonal}, not against random directions. Under the baseline-corrected comparison (norm-matched random directions in the same eigenbasis), concept directions are more tail-aligned than their matched nulls in 13 of 17 models, with a mean true-minus-null Gini gap of $-0.082$; Llama~3.2 (1B and 3B), Mistral-7B-v0.3, and OLMoE are the exceptions (\Cref{app:robustness-controls}).

\paragraph{Generalization to a broader semantic set.}
Because only 7 of the 27 concepts are semantic, we tested whether the effect depends on this narrow selection by adding 15 further BATS semantic categories (encyclopedic and lexicographic: country--language, name--nationality/occupation, animal--young/sound/shelter, hypernyms, hyponyms, meronyms, synonyms, and antonyms), more than tripling the semantic set from 7 to 22, and recomputing anti-concentration across all 17 models. The signature is unchanged: mean expanded-semantic Gini $=-0.278$, statistically identical to the original semantic subset ($-0.278$; Wilcoxon signed-rank $p=0.71$), with all 17 models negative (one-sample $t=-12.3$, $p=1.5\times10^{-9}$; model as the unit of analysis). The effect is therefore not an artifact of the original concept selection. Full per-model results, an instruction-tuned-model replication, and a centered-covariance control are in \Cref{app:expanded-semantic}.

\paragraph{Robustness to spectral design choices.}
Anti-concentration does not depend on how $\Sigma$ is constructed. \Cref{tab:sensitivity} summarizes the mean Gini deviation across eight covariance constructions spanning the token set (full vocabulary, language subset, input embeddings), centering, and frequency weighting: every variant is negative and significant in 16--17 of the 17 models. The effect is likewise invariant to the spectral cutoff ($k\in\{5\%,10\%,20\%\}$; \Cref{app:pos-sensitivity}) and emerges through depth---near-neutral at the embedding layer (mean Gini $-0.009$) and tail-aligned from the early transformer layers onward ($-0.17$ at layer~0, deepening to $-0.30$ near mid-depth; \Cref{app:robustness-controls}).

\begin{table}[t]
\centering\small
\caption{Sensitivity of anti-concentration (mean Gini deviation over 17 models) to covariance construction. All variants are negative; the last column is the number of models with $p<0.05$. The default row ($-0.275$) is computed from the saved language-subset covariance of this sensitivity sweep and differs slightly from the $-0.282$ headline mean in \Cref{sec:anti-concentration}, which comes from the main experiment-1 run; the two pipelines agree in sign and magnitude on every model.}
\label{tab:sensitivity}
\resizebox{\columnwidth}{!}{%
\begin{tabular}{lcc}
\toprule
\textbf{Covariance construction} & \textbf{Mean Gini} & \textbf{Sig.} \\
\midrule
Language subset, uncentered (default)   & $-0.275$ & 17/17 \\
Language subset, centered                & $-0.288$ & 17/17 \\
Language subset, freq.-weighted          & $-0.350$ & 17/17 \\
\quad + centered                         & $-0.367$ & 17/17 \\
Full vocabulary, uncentered              & $-0.274$ & 17/17 \\
Full vocabulary, centered                & $-0.271$ & 16/17 \\
Input embeddings, uncentered             & $-0.271$ & 17/17 \\
Input embeddings, centered               & $-0.269$ & 17/17 \\
\bottomrule
\end{tabular}%
}
\end{table}

\begin{table}[t]
\centering
\small
\caption{Representative spectral metrics. Concept \scm exceeds the random baseline on average, and all Gini deviations are negative across all 17 models; the full table is in \Cref{app:full-tables}.}
\label{tab:scm-summary}
\resizebox{\columnwidth}{!}{%
\begin{tabular}{lccccc}
\toprule
\textbf{Model} & \textbf{$d$} & \textbf{Conc.\ \scm} & \textbf{Rand.\ \scm} & \textbf{Gap} & \textbf{Gini} \\
\midrule
Qwen2.5-0.5B  & 896  & 0.997 & 0.711 & $+$0.286 & $-$0.356 \\
Qwen2.5-14B   & 5120 & 0.996 & 0.790 & $+$0.205 & $-$0.346 \\
Qwen3-8B      & 4096 & 1.000 & 0.783 & $+$0.217 & $-$0.359 \\
Llama-3.2-3B  & 3072 & 0.771 & 0.771 & $+$0.001 & $-$0.159 \\
Gemma-2-9b    & 3584 & 0.927 & 0.849 & $+$0.078 & $-$0.364 \\
Mistral-7B    & 4096 & 0.908 & 0.792 & $+$0.116 & $-$0.199 \\
JetMoE-8B     & 2048 & 0.945 & 0.741 & $+$0.205 & $-$0.242 \\
\midrule
\textbf{Mean (all 17)} & & 0.926 & 0.758 & $+$0.169 & $-$0.282 \\
\bottomrule
\end{tabular}
}
\end{table}

\Cref{tab:scm-summary} highlights two patterns visible in the full data. First, all 17 models show negative Gini, supporting broad anti-concentration across families and scales. Second, the \scm gap is positive on average but small or slightly negative in the Llama~3.2 and OLMoE cases; these models still anti-concentrate by the Gini metric---two complementary measures that disagree when the concept-energy CDF lies below uniform but with a center-of-mass close to the random baseline.

\paragraph{Anti-concentration is not just anisotropy.}
Anisotropy refers to the well-documented observation that transformer activations cluster in a narrow cone, with a few dimensions carrying disproportionate variance~\citep{ethayarajh2019contextual,gao2019representation,timkey2021all}. Our claim is distinct on three counts. (i)~Anti-concentration is a property of \emph{specific concept directions} rather than of the spectrum itself. Anisotropy alone does move the Gini deviation: random unit vectors evaluated in these spectra are themselves negative ($-0.16$ to $-0.28$), so the raw statistic cannot separate the two. The separating test is the norm-matched random-direction null reported above, which subtracts the anisotropic component and still leaves concept directions further into the tail in 13 of 17 models. (ii)~The correlation between random-vector \scm and concept-vector \scm is negligible across our 17 models ($r=0.019$, $p=0.94$); concepts do not anti-concentrate \emph{because} the spectrum is concentrated, but in addition to it. (iii)~The \emph{dual geometry} (\Cref{sec:dual-geometry}) is incompatible with a pure anisotropy account: concept-specific unembedding-row differences \emph{concentrate} in the same eigenbasis where activation-space concepts anti-concentrate. If anisotropy alone drove the effect, both views of the concept would inherit the same spectral bias.

\section{Ruling Out the Subtraction Artifact}
\label{sec:artifact-check}

The \dom estimator mechanically cancels components shared between paired tokens, potentially biasing it toward low-variance directions. We rule this out with two methods that do not involve subtraction.

\paragraph{SAE features.}
Top-$k$ differentially activating features on Gemma Scope (gemma-2-2b layer 20, gemma-2-9b layer 31, width 16K) and Llama Scope (Llama-3.1-8B layer 24, width 32K) all show strong anti-concentration (\Cref{tab:sae-check}). Top-5 selection \emph{strengthens} the effect relative to top-1 on gemma-2-2b, and feature collapse does not drive the result: the most-collapsed feature (used by 5/27 concepts) was the \emph{weakest} contributor, and removing every concept that touches a collapsed feature ($N=18$ remaining) strengthened the aggregate (\Cref{app:sae-collapse}).

\begin{table}[t]
\centering
\small
\caption{SAE artifact check: anti-concentration is highly significant across three models in two families with no subtraction involved.}
\label{tab:sae-check}
\resizebox{\columnwidth}{!}{%
\begin{tabular}{lccc}
\toprule
\textbf{Model (SAE)} & \textbf{Mean Gini} & \textbf{$t$} & \textbf{$p$} \\
\midrule
Gemma-2-2b (top-1)     & $-0.230$ & $-23.6$ & $4.5\times 10^{-19}$ \\
Gemma-2-2b (top-5)     & $-0.240$ & $-51.3$ & $1.2\times 10^{-27}$ \\
Gemma-2-9b (top-5)     & $-0.276$ & $-52.8$ & $5.6\times 10^{-28}$ \\
Llama-3.1-8B (top-5)   & $-0.193$ & $-49.1$ & $3.6\times 10^{-27}$ \\
\bottomrule
\end{tabular}
}
\end{table}

\begin{figure}[t]
  \centering
  \includegraphics[width=\columnwidth]{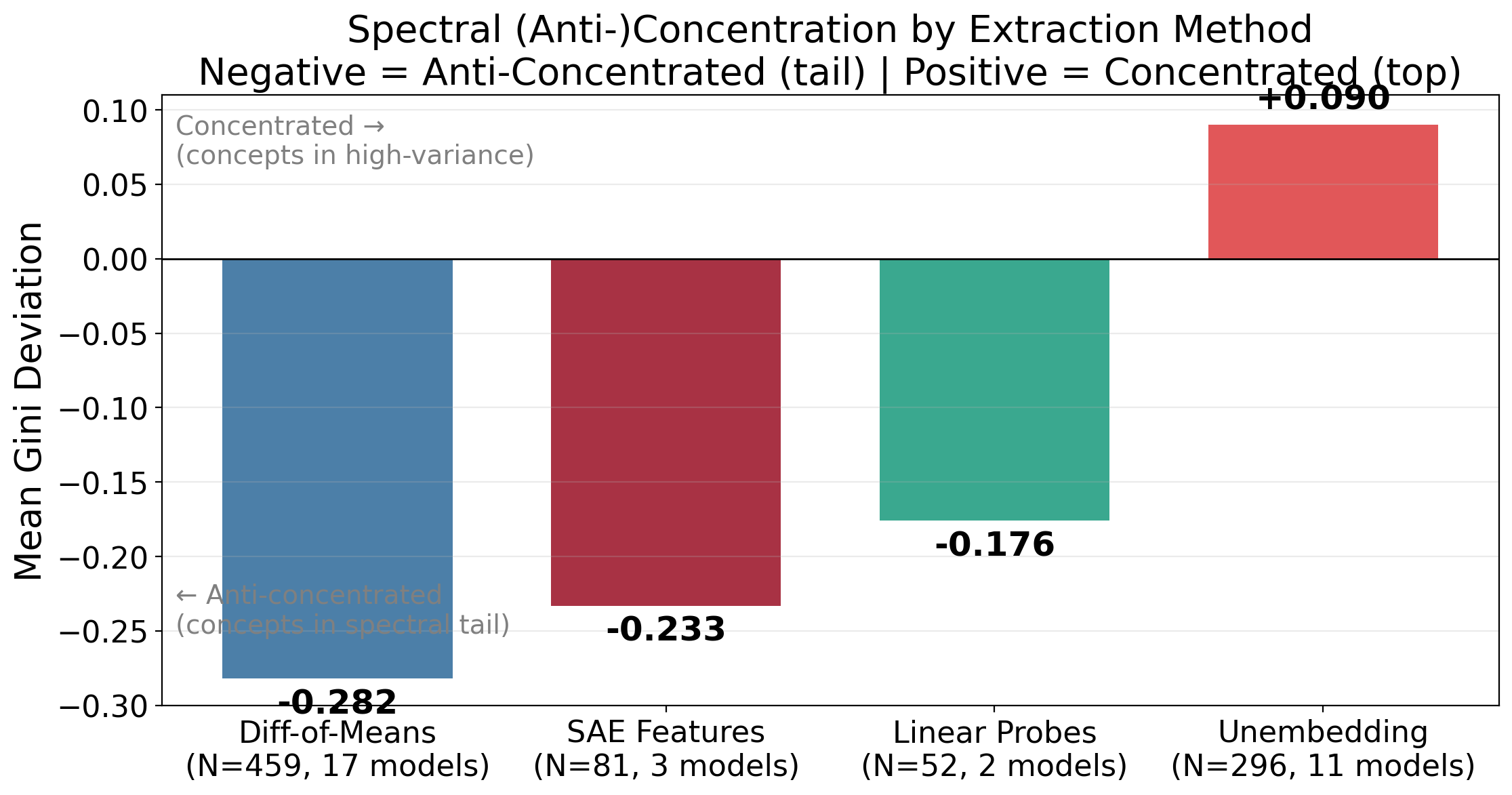}
  \caption{Cross-method Gini deviation across models. \dom, SAE features, and linear probes all yield negative Gini (anti-concentration); unembedding-row contrasts have the opposite sign---the dual geometry of \Cref{sec:dual-geometry}.}
  \label{fig:method-comparison}
\end{figure}

\paragraph{Linear probes.}
L2-regularized logistic regression on uncentered activations reaches mean accuracy $>0.99$ on Llama-3.2-3B (Gini $=-0.167$, $p<10^{-13}$) and Qwen2.5-3B ($-0.185$, $p<10^{-15}$). Probes failed on gemma-2-9b (21 of 26 concepts at exactly chance accuracy, and all 26 below $0.53$), likely due to RMSNorm interaction with uncentered inputs; we exclude this model from probe-based claims (SAE evidence on the same model is unaffected). All three activation-space methods yield negative Gini (\Cref{fig:method-comparison}).

\section{The Dual Geometry}
\label{sec:dual-geometry}

\paragraph{Unembedding contrasts concentrate.}
Constructing concept directions directly from static unembedding rows, $v_\text{unembed}=\text{mean}_i(\gamma(w_i^+)-\gamma(w_i^-))$ over single-token pairs, the resulting directions \emph{concentrate} in high-variance eigendirections (positive Gini), the opposite sign of activation-space directions.

\begin{figure*}[t]
  \centering
  \includegraphics[width=0.92\textwidth]{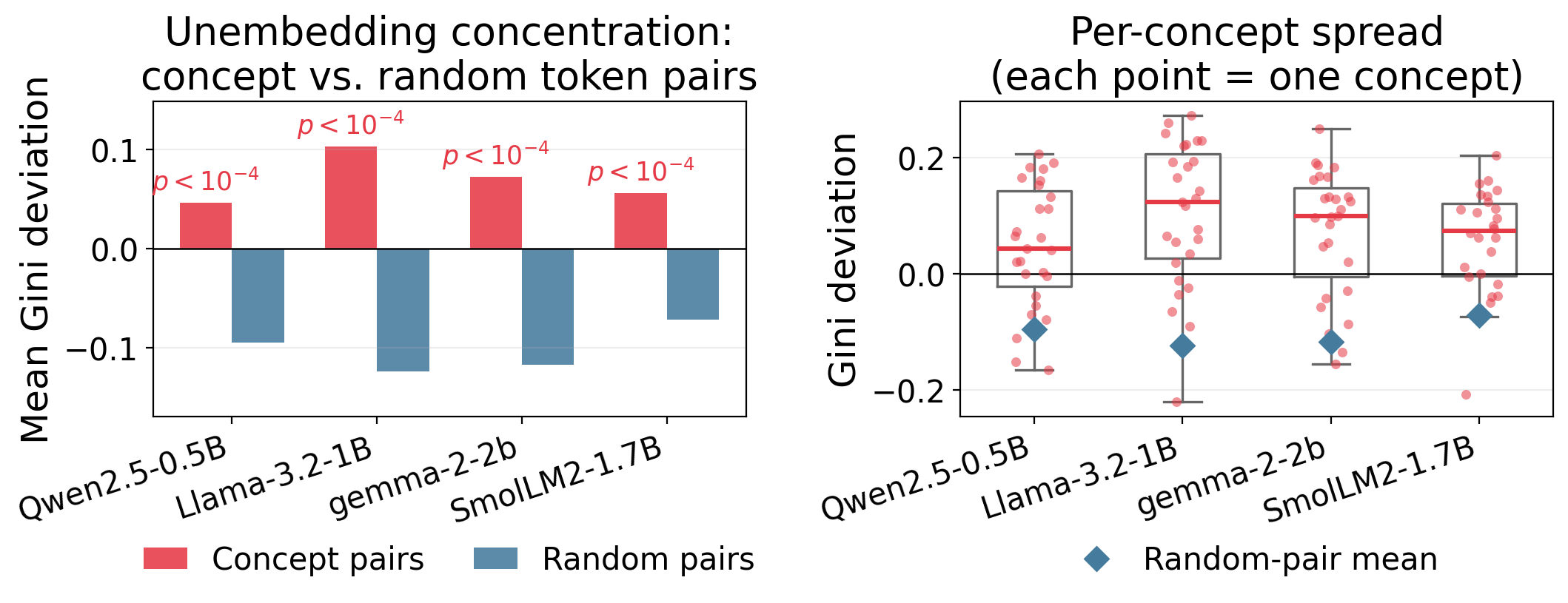}
  \caption{Random unembedding null. \textbf{Left:} random token-pair differences (blue) mildly anti-concentrate, while concept-specific pairs (red) actively concentrate in the top eigenvalues, on all four models ($p<10^{-4}$, 1{,}000 random pairs per model). \textbf{Right:} the per-concept spread behind those means---each point is one concept, with the random-pair mean marked as a blue diamond---showing the effect is carried by the bulk of the concept set rather than a few outliers.}
  \label{fig:unembed-null}
\end{figure*}

\paragraph{Concept pairs are non-random.}
Against a null of 1{,}000 random token-pair differences from the same vocabulary, concept pairs concentrate significantly more (\Cref{fig:unembed-null}). Random pairs mildly anti-concentrate (mean Gini $\approx-0.10$); concept pairs concentrate (mean $\approx+0.07$); $p<10^{-4}$ on Qwen2.5-0.5B, Llama-3.2-1B, Gemma-2-2b, and SmolLM2-1.7B.

\paragraph{Interpretation.}
The unembedding matrix places \emph{concept contrasts along high-variance directions}, where they maximally influence logit differences. Activation-space representations encode the same concepts in the spectral tail, away from the dominant structural variance. Additional layer-wise controls show that this rotation emerges during contextualized processing rather than being inherited wholesale from static embeddings: mean Gini moves from near-neutral at the embedding layer ($-0.009$) to tail-aligned at layer~0 ($-0.171$) and the final layer ($-0.277$; \Cref{app:robustness-controls}).

\section{Why Concepts Anti-Concentrate}
\label{sec:why}

The dual geometry motivates two functional tests: a steering experiment isolating spectral components, and a syntactic probing experiment characterizing what each subspace encodes.

\paragraph{Split-injection interference.}
For each concept we project onto the top-$k$ (\emph{shouting}) and bottom-$k$ (\emph{whispering}) eigenvectors ($k=0.1d$) and steer at four mid-depth layers (50/62/75/87\%) at scaling $\alpha$, measuring perplexity on 10 held-out Wikipedia passages across five models. Large $\alpha$ destroys both arms---on Mistral-7B at $\alpha=5$, shouting and whispering raise perplexity by $+18$K\% and $+59$K\%---and comparing two collapsed models does not measure interference. We therefore impose a \emph{validity window}: a configuration counts only if \emph{both} arms stay below a $+1000\%$ perplexity increase. The criterion concerns measurement, not the hypothesis; it reads the perplexity magnitudes alone, independently of the sign or significance of the asymmetry. Thirteen of the 30 configurations qualify (\Cref{tab:split-injection-full}).

\begin{figure*}[t]
  \centering
  \includegraphics[width=0.95\textwidth]{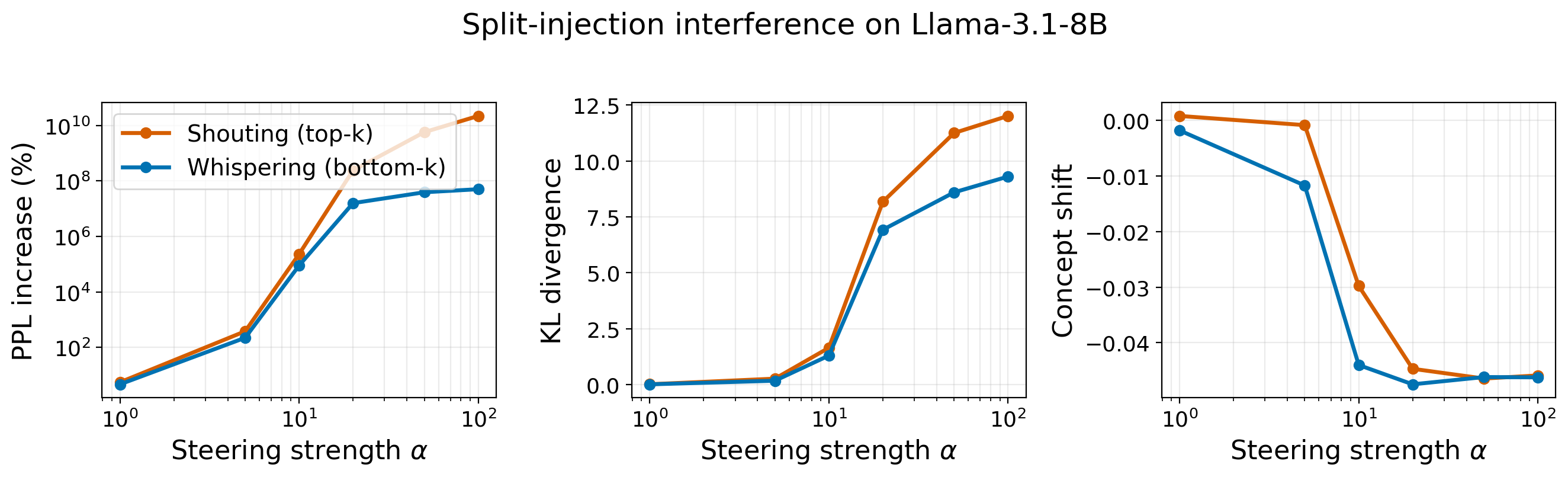}
  \caption{Split-injection on Llama-3.1-8B. PPL increase (left), KL divergence (center), and concept shift (right) vs.\ steering strength $\alpha$. At $\alpha=5$, the shouting component produces $+383\%$ PPL increase vs.\ $+223\%$ for whispering (paired Cohen's $d_z=1.09$, $p<10^{-4}$; inside the validity window of \S\ref{sec:why}).}
  \label{fig:split-injection-llama-main}
\end{figure*}

Inside the window the picture is consistent (\Cref{tab:split-injection,fig:split-injection-llama-main}): seven configurations reach significance and \emph{all seven} favor shouting, spanning four of five models (gemma-2-2b, both Llamas, Qwen2.5-3B); Mistral-7B has one in-window configuration, not significant. No significant reversal occurs inside the window. The sweep contains exactly one \emph{significant} reversal---Qwen2.5-3B at $\alpha=20$ ($d_z=-0.43$, $p=.035$)---and it falls outside the window, its whispering arm having raised perplexity by $+197$K\%. Two in-window rows do carry a negative $d_z$ (gemma-2-2b at $\alpha=1$, Qwen2.5-3B at $\alpha=10$), but both are far from significance ($p=.15$ and $p=.51$). A stricter $+100\%$ window retains five significant configurations, all positive.

Two caveats. The magnitude stays $\alpha$-dependent within the window---$d_z$ on gemma-2-2b runs from $-0.29$ at $\alpha=1$ (not significant) to $+1.19$ at $\alpha=20$---so an effect size at one $\alpha$ is not a model-level constant. And the window is post-hoc, applied to a completed sweep; we state it explicitly so the excluded configurations stay visible in \Cref{tab:split-injection-full}. The seven configurations are also nested within four models, and the 30 sweep $p$-values are uncorrected, so the model-level count (4 of 5) is the conservative reading.

\begin{table}[t]
\centering
\small
\caption{Split-injection interference asymmetry at the strongest in-window configuration per model (both arms below a $+1000\%$ perplexity increase; see text). $d_z$ is the paired Cohen's $d$ over the 27 per-concept shout$-$whisper differences, matching the paired $t$-test $p$; positive means shouting disrupts more. All four significant rows are positive, as are the three further significant in-window configurations not shown (gemma-2-2b $\alpha=10$, Llama-3.2-3B $\alpha=5$, Qwen2.5-3B $\alpha=1$)---seven of the 13 in-window configurations reach significance, all favoring shouting. Full sweep: \Cref{tab:split-injection-full}.}
\label{tab:split-injection}
\resizebox{\columnwidth}{!}{%
\begin{tabular}{llccc}
\toprule
\textbf{Model} & \textbf{$\alpha$} & \textbf{Shout/Whisper} & \textbf{$d_z$} & \textbf{$p$} \\
\midrule
gemma-2-2b     & 20 & $+4.8\%$ / $+1.8\%$    & $+1.19$ & $<\!10^{-4}$ \\
Llama-3.1-8B   & 5  & $+383\%$ / $+223\%$    & $+1.09$ & $<\!10^{-4}$ \\
Llama-3.2-3B   & 1  & $+4.6\%$ / $+2.9\%$    & $+0.90$ & $.0001$ \\
Qwen2.5-3B     & 5  & $+2.4\%$ / $+1.5\%$    & $+0.60$ & $.0045$ \\
\midrule
\multicolumn{5}{l}{\textit{No significant in-window asymmetry:}} \\
Mistral-7B     & 1  & $+5.6\%$ / $+5.1\%$    & $+0.31$ & $.12$ \\
\bottomrule
\end{tabular}
}
\end{table}

\paragraph{POS-tag probing.}
We train 13-class logistic regression classifiers on activations from the first 500 UD English-EWT sentences~\citep{nivre2016universal} (6{,}585 words after excluding punctuation/symbols and POS tags with $<30$ samples), projected onto the top-$k$ and bottom-$k$ eigenvectors of $\Sigma$ ($k=0.1d$). \Cref{tab:pos-probe} shows that 6/8 models place significantly more POS information in the high-variance subspace; both Qwen~2.5 models reverse this. Notably, Qwen~3-4B---same vocabulary, different training---shows the expected positive gap.

\begin{table}[t]
\centering
\small
\caption{POS-tag probing accuracy on top-$k$ (high-variance) vs.\ bottom-$k$ (low-variance) subspaces at $k=0.1d$. Six of eight models show top-$k$ $>$ bottom-$k$; both Qwen~2.5 models reverse. 95\% bootstrap CIs and the full table are in \Cref{app:pos-probe}.}
\label{tab:pos-probe}
\begin{tabular}{lccc}
\toprule
\textbf{Model} & \textbf{Top-$k$} & \textbf{Bot-$k$} & \textbf{Gap ($p$)} \\
\midrule
gemma-2-2b     & .554 & .506 & $+.048$ (.0001) \\
Llama-3.2-3B   & .628 & .576 & $+.052$ (.0001) \\
Llama-3.1-8B   & .641 & .625 & $+.016$ (.012)  \\
Mistral-7B     & .624 & .607 & $+.017$ (.013)  \\
SmolLM2-1.7B   & .560 & .536 & $+.024$ (.002)  \\
Qwen3-4B       & .572 & .555 & $+.018$ (.001)  \\
\midrule
Qwen2.5-3B     & .557 & .576 & $-.019$ ($<\!.0001$) \\
Qwen2.5-7B     & .572 & .586 & $-.014$ (.002)  \\
\bottomrule
\end{tabular}
\end{table}

A UD Chinese-GSD control on Qwen2.5-3B and Qwen~3-4B yields bottom-$k$ $>$ top-$k$ for both, so the Qwen~2.5 reversal cannot be reduced to a single English-vs-multilingual mismatch (\Cref{app:pos-probe}). The qualitative pattern is invariant under $k\in\{5\%,10\%,20\%\}$ (\Cref{app:pos-sensitivity}). Both subspaces underperform full-dimensional probes (78--86\%); the high-variance subspace carries a measurable \emph{bias} toward syntactic information.

\paragraph{Scope of the functional claim.}
Anti-concentration itself replicates across all 17 models and three extraction methods; the \emph{functional} interpretation---that high-variance directions disproportionately carry syntactic load and that injecting along them disrupts fluency more---is supported by both functional probes but remains partial. The two functional probes diverge informatively. For \emph{steering}, once the validity window is imposed, every significant in-window asymmetry is positive, across four of five models including Qwen2.5-3B; only Mistral-7B is undetermined. For \emph{POS probing}, which has no steering strength and hence no window, the Qwen~2.5 reversal stands and is not explained by the Chinese control. The residual disagreement is thus specific to Qwen~2.5's syntactic probing rather than a general failure of the account. Steering or probing recommendations derived from it should be evaluated per-model.

\section{Discussion and Conclusion}
\label{sec:discussion}

\paragraph{Connection to superposition.}
If the high-variance eigenspace is claimed by frequent, high-information syntactic regularities, the effective dimensionality available for semantic encoding is $d_\text{eff}\ll d$, forcing superposition for semantic features~\citep{elhage2022toy,scherlis2022polysemanticity} even when $d$ is large. Anti-concentration is thus a candidate spectral signature of how transformers allocate representational capacity: syntax claims the ``loud'' subspace where it can robustly survive residual-stream noise, while semantic concepts share the ``quiet'' subspace via superposition.

\paragraph{Scaling and architecture.}
The Qwen~2.5 (0.5B--14B) and Qwen~3 (0.6B--8B) families let us check whether anti-concentration changes with capacity. The \scm gap remains positive at all scales (Qwen~2.5: $+0.205$ to $+0.286$; Qwen~3: $+0.215$ to $+0.259$) without a clear monotonic trend (\Cref{app:scaling}); MoE models (OLMoE, JetMoE) show the same pattern as dense families, and Gemma's condition number---the largest after Mistral's---does not abolish the effect (\Cref{app:architecture}). We interpret this as a property present at all tested scales rather than one that emerges or vanishes with capacity.

\paragraph{Implications for interpretability and steering.}
PCA-based visualizations and low-rank approximations risk missing concept-level structure; interpretability tools should attend to the spectral tail. On models where the interference asymmetry appears, it suggests a practical refinement to activation engineering~\citep{turner2024activation,panickssery2024steering}: projecting steering vectors into the low-variance subspace before injection may reduce fluency disruption, though whether it preserves concept-manipulation strength is untested here; this may also explain why naive steering can cause aphasia~\citep{arditi2024refusal}. The complementary finding that whitening improves sentence embeddings~\citep{su2021whitening,huang2021whiteningbert} is consistent with this picture: whitening amplifies the low-variance directions where concepts reside. \citet{sharma2023truth} report reasoning gains from removing top SVD components (LASER), again consistent with the dual geometry we describe.

\paragraph{The Qwen~2.5 anomaly.}
Qwen~2.5 reverses the POS pattern while Qwen~3 does not, despite sharing vocabulary, and the reversal persists at $k\in\{5\%,10\%,20\%\}$. Chinese probing on UD Chinese-GSD reverses for \emph{both} Qwen families, so language alone does not explain the English-vs-rest split. A natural hypothesis is that Qwen~2.5's relatively flat eigenspectrum (the smallest top-10\% variance capture in our sample, 26--27\%) makes any single cutoff insufficient to separate syntactic from semantic directions; the $k$-sensitivity analysis (\Cref{app:pos-sensitivity}) argues against this, however, since the reversal persists at $k=20\%$. We hypothesize that training-pipeline differences distribute syntactic information differently across the eigenspectrum, and leave a controlled investigation of pre-training mixture effects to future work.

\paragraph{Conclusion.}
We set out to validate the causal inner product as a basis for cross-lingual transport; matched-spectrum randomization shows no advantage for the specific eigendirections of $\Sigma$ over randomized directions with the same spectrum, suggesting \wca acts as spectral regularization. The failure exposed a broader pattern: word-level concept directions---across SAE features, linear probes, and \dom vectors over a 17-model core suite---systematically avoid high-variance eigendirections, while static unembedding-row contrasts do the opposite. Split-injection and POS-tag probing give \emph{partial} functional support: restricted to steering strengths at which both arms remain interpretable, every significant interference asymmetry favors the high-variance component, across four of five models; POS probing agrees in six of eight. Thus \emph{concepts whisper}: word-level semantics occupy the spectrally quiet subspace, a robust effect across 17 models and three extraction methods; the complementary claim that the high-variance subspace disproportionately carries syntactic load is supported but not settled: the steering evidence is uniform in sign once uninterpretable configurations are excluded, yet rests on a post-hoc validity window, and POS probing still reverses on Qwen~2.5. We present it as a candidate mechanism rather than a rule. Either way, the geometry has immediate consequences for how we probe, interpret, and steer large language models.

\section*{Limitations}
\label{sec:limitations}

Our concepts are word-level counterfactual pairs~\citep{park2024linear,gladkova2016analogy}; whether anti-concentration extends to abstract concepts such as honesty, sarcasm, or planning is open. We use an unembedding second moment from a single matrix; centered, frequency-weighted, and input-embedding controls preserve the main effect (\Cref{app:robustness-controls}), but activation-covariance spectra may reveal additional structure. Our layer-wise analysis is descriptive; a causal per-layer intervention study of where the vocabulary-to-reasoning rotation is implemented remains future work.

Split-injection steers at four layers simultaneously; per-layer interference profiles were not logged, so we cannot localize the asymmetry in depth. Each perplexity estimate rests on 10 held-out Wikipedia passages, so per-configuration estimates are noisy. The steering effect size is also $\alpha$-dependent: it is not a single number per model, and at large $\alpha$ both spectral components drive perplexity up by orders of magnitude, at which point a comparison between them no longer measures interference. We therefore read the asymmetry only inside a stated validity window (both arms below a $+1000\%$ perplexity increase), which admits 13 of our 30 configurations; the window is a post-hoc criterion applied to a completed sweep rather than a pre-registered one, and a study designed around it would choose $\alpha$ per model in advance. Inside the window the asymmetry is positive wherever significant and never significantly reversed, but Mistral-7B contributes only one in-window configuration, so its status is genuinely undetermined rather than null. POS-probing significance is based on 5 cross-validation folds plus token-level bootstrap; sentence- or document-clustered resampling would be stronger. The high-variance/syntax association is also established on English UD data only: on UD Chinese-GSD the top-minus-bottom gap reverses in both models tested.

The core suite is decoder-only \emph{base} (non-instruct) variants, chosen to isolate pre-training geometry from post-training alignment effects. Two instruction-tuned checkpoints tested as a control show essentially unchanged anti-concentration (\Cref{app:expanded-semantic}), but two models are not a systematic test, so whether RLHF or instruction tuning alters the spectral allocation of concept information remains open. Encoder or encoder-decoder architectures may also differ.

\section*{Ethical Considerations}
\label{sec:ethics}

This work uses only publicly available pre-trained language models and publicly available linguistic datasets (UD English-EWT and UD Chinese-GSD~\citep{nivre2016universal}); no human subjects were involved and no personally identifiable information was used. The phenomena we report---spectral structure and steering interference---are properties of model internals rather than user-facing outputs; we do not introduce new capabilities for generating harmful content. Our split-injection results may inform safer steering by suggesting that interventions targeted at low-variance directions could reduce fluency disruption, but they also make explicit which subspaces are most sensitive to manipulation, which could in principle be misused; we believe the interpretability and safety benefits clearly outweigh this risk. Dataset statistics and the compute budget are reported in \Cref{app:hyperparams} and, together with AI-assistance disclosures, in the Responsible NLP Checklist. All artifacts are used consistently with their licenses: UD English-EWT and UD Chinese-GSD under CC BY-SA 4.0; Wikipedia text under CC BY-SA; and the models under Apache-2.0 (Mistral-7B-v0.3, SmolLM2, OLMoE, JetMoE, and most Qwen releases), the Llama Community License (Llama~3.x), the Gemma Terms of Use (Gemma~2), and the Qwen Research License for Qwen2.5-3B (all other Qwen2.5 and Qwen3 checkpoints we use are Apache-2.0).

\section*{Acknowledgments}
We thank Pukar Hamal and Laxman M.\ Basnet of SecurityPal AI for their support and guidance, and for enabling us to pursue exploratory research beyond our day-to-day work. We also thank the SecurityPal AI Data Science team for their discussions, feedback, and support throughout this research.

\bibliography{references}

\begin{thebibliography}{46}
\providecommand{\natexlab}[1]{#1}

\bibitem[{Arditi et~al.(2024)Arditi, Obeso, Syed, Paleka, Panickssery, Gurnee,
  and Nanda}]{arditi2024refusal}
Andy Arditi, Oscar Obeso, Aaquib Syed, Daniel Paleka, Nina Panickssery, Wes
  Gurnee, and Neel Nanda. 2024.
\newblock \href {https://arxiv.org/abs/2406.11717} {Refusal in language models
  is mediated by a single direction}.
\newblock \emph{arXiv preprint arXiv:2406.11717}.

\bibitem[{Bricken et~al.(2023)Bricken, Templeton, Batson, Chen, Jermyn,
  Conerly, Turner et~al.}]{bricken2023towards}
Trenton Bricken, Adly Templeton, Joshua Batson, Brian Chen, Adam Jermyn, Tom
  Conerly, Nick Turner, and 1 others. 2023.
\newblock \href
  {https://transformer-circuits.pub/2023/monosemantic-features/index.html}
  {Towards monosemanticity: Decomposing language models with dictionary
  learning}.
\newblock \emph{Transformer Circuits Thread}.

\bibitem[{Burns et~al.(2023)Burns, Ye, Klein, and
  Steinhardt}]{burns2023discovering}
Collin Burns, Haotian Ye, Dan Klein, and Jacob Steinhardt. 2023.
\newblock \href {https://arxiv.org/abs/2212.03827} {Discovering latent
  knowledge in language models without supervision}.
\newblock In \emph{International Conference on Learning Representations
  (ICLR)}.

\bibitem[{Conneau et~al.(2018)Conneau, Lample, Ranzato, Denoyer, and
  J{\'e}gou}]{conneau2018word}
Alexis Conneau, Guillaume Lample, Marc'Aurelio Ranzato, Ludovic Denoyer, and
  Herv{\'e} J{\'e}gou. 2018.
\newblock \href {https://arxiv.org/abs/1710.04087} {Word translation without
  parallel data}.
\newblock In \emph{International Conference on Learning Representations
  (ICLR)}.

\bibitem[{Conneau et~al.(2020)Conneau, Wu, Li, Zettlemoyer, and
  Stoyanov}]{conneau2020emerging}
Alexis Conneau, Shijie Wu, Haoran Li, Luke Zettlemoyer, and Veselin Stoyanov.
  2020.
\newblock \href {https://doi.org/10.18653/v1/2020.acl-main.536} {Emerging
  cross-lingual structure in pretrained language models}.
\newblock In \emph{Annual Meeting of the Association for Computational
  Linguistics (ACL)}, pages 6022--6034.

\bibitem[{Cunningham et~al.(2024)Cunningham, Ewart, Riggs, Huben, and
  Sharkey}]{cunningham2024sparse}
Hoagy Cunningham, Aidan Ewart, Logan Riggs, Robert Huben, and Lee Sharkey.
  2024.
\newblock \href {https://arxiv.org/abs/2309.08600} {Sparse autoencoders find
  highly interpretable features in language models}.
\newblock In \emph{International Conference on Learning Representations
  (ICLR)}.

\bibitem[{Elhage et~al.(2022)Elhage, Hume, Olsson, Schiefer, Henighan, Kravec,
  Hatfield-Dodds, Lasenby, Drain, Chen, Grosse, McCandlish, Kaplan, Amodei,
  Wattenberg, and Olah}]{elhage2022toy}
Nelson Elhage, Tristan Hume, Catherine Olsson, Nicholas Schiefer, Tom Henighan,
  Shauna Kravec, Zac Hatfield-Dodds, Robert Lasenby, Dawn Drain, Carol Chen,
  Roger Grosse, Sam McCandlish, Jared Kaplan, Dario Amodei, Martin Wattenberg,
  and Christopher Olah. 2022.
\newblock \href {https://arxiv.org/abs/2209.10652} {Toy models of
  superposition}.
\newblock \emph{Transformer Circuits Thread}.

\bibitem[{Engels et~al.(2025)Engels, Michaud, Liao, Gurnee, and
  Tegmark}]{engels2025not}
Joshua Engels, Eric~J. Michaud, Isaac Liao, Wes Gurnee, and Max Tegmark. 2025.
\newblock \href {https://arxiv.org/abs/2405.14860} {Not all language model
  features are one-dimensionally linear}.
\newblock In \emph{International Conference on Learning Representations
  (ICLR)}.

\bibitem[{Ethayarajh(2019)}]{ethayarajh2019contextual}
Kawin Ethayarajh. 2019.
\newblock \href {https://doi.org/10.18653/v1/D19-1006} {How contextual are
  contextualized word representations? {C}omparing the geometry of {BERT},
  {ELMo}, and {GPT-2} embeddings}.
\newblock In \emph{Conference on Empirical Methods in Natural Language
  Processing (EMNLP)}, pages 55--65.

\bibitem[{Gao et~al.(2019)Gao, He, Tan, Qin, Wang, and
  Liu}]{gao2019representation}
Jun Gao, Di~He, Xu~Tan, Tao Qin, Liwei Wang, and Tie-Yan Liu. 2019.
\newblock \href {https://arxiv.org/abs/1907.12009} {Representation degeneration
  problem in training natural language generation models}.
\newblock In \emph{International Conference on Learning Representations
  (ICLR)}.

\bibitem[{Gao et~al.(2025)Gao, Dupr{\'e}~la Tour, Tillman, Goh, Troll, Radford,
  Sutskever, Leike, and Wu}]{gao2025scaling}
Leo Gao, Tom Dupr{\'e}~la Tour, Henk Tillman, Gabriel Goh, Rajan Troll, Alec
  Radford, Ilya Sutskever, Jan Leike, and Jeffrey Wu. 2025.
\newblock \href {https://arxiv.org/abs/2406.04093} {Scaling and evaluating
  sparse autoencoders}.
\newblock In \emph{International Conference on Learning Representations
  (ICLR)}.

\bibitem[{Gladkova et~al.(2016)Gladkova, Drozd, and
  Matsuoka}]{gladkova2016analogy}
Anna Gladkova, Aleksandr Drozd, and Satoshi Matsuoka. 2016.
\newblock \href {https://doi.org/10.18653/v1/N16-2002} {Analogy-based detection
  of morphological and semantic relations with word embeddings: What works and
  what doesn't}.
\newblock \emph{Proceedings of the NAACL Student Research Workshop}, pages
  8--15.

\bibitem[{Gurnee and Tegmark(2023)}]{gurnee2023language}
Wes Gurnee and Max Tegmark. 2023.
\newblock \href {https://arxiv.org/abs/2310.02207} {Language models represent
  space and time}.
\newblock \emph{arXiv preprint arXiv:2310.02207}.

\bibitem[{He et~al.(2024)He, Shu, Ge, Chen, Wang, Zhou, Liu, Guo, Huang, Wu,
  Jiang, and Qiu}]{he2024llama}
Zhengfu He, Wentao Shu, Xuyang Ge, Lingjie Chen, Junxuan Wang, Yunhua Zhou,
  Frances Liu, Qipeng Guo, Xuanjing Huang, Zuxuan Wu, Yu-Gang Jiang, and Xipeng
  Qiu. 2024.
\newblock \href {https://arxiv.org/abs/2410.20526} {Llama scope: Extracting
  millions of features from {Llama}-3.1-8{B} with sparse autoencoders}.
\newblock \emph{arXiv preprint arXiv:2410.20526}.

\bibitem[{Huang et~al.(2021)Huang, Tang, Zhong, Lu, Shou, Gong, Jiang, and
  Duan}]{huang2021whiteningbert}
Junjie Huang, Duyu Tang, Wanjun Zhong, Shuai Lu, Linjun Shou, Ming Gong, Daxin
  Jiang, and Nan Duan. 2021.
\newblock \href {https://doi.org/10.18653/v1/2021.findings-emnlp.23}
  {Whitening{BERT}: An easy unsupervised sentence embedding approach}.
\newblock In \emph{Findings of EMNLP}, pages 238--244.

\bibitem[{Huh et~al.(2024)Huh, Cheung, Wang, and Isola}]{huh2024platonic}
Minyoung Huh, Brian Cheung, Tongzhou Wang, and Phillip Isola. 2024.
\newblock \href {https://arxiv.org/abs/2405.07987} {Position: The platonic
  representation hypothesis}.
\newblock In \emph{International Conference on Machine Learning (ICML)}, volume
  235 of \emph{PMLR}, pages 20617--20642.

\bibitem[{Kovaleva et~al.(2021)Kovaleva, Kulshreshtha, Rogers, and
  Rumshisky}]{kovaleva2021bert}
Olga Kovaleva, Saurabh Kulshreshtha, Anna Rogers, and Anna Rumshisky. 2021.
\newblock \href {https://doi.org/10.18653/v1/2021.findings-acl.300} {{BERT}
  busters: Outlier dimensions that disrupt transformers}.
\newblock In \emph{Findings of ACL-IJCNLP}, pages 3392--3405.

\bibitem[{Lan et~al.(2024)Lan, Torr, Meek, Khakzar, Krueger, and
  Barez}]{lan2024sparse}
Michael Lan, Philip Torr, Austin Meek, Ashkan Khakzar, David Krueger, and Fazl
  Barez. 2024.
\newblock \href {https://arxiv.org/abs/2410.06981} {Sparse autoencoders reveal
  universal feature spaces across large language models}.
\newblock \emph{arXiv preprint arXiv:2410.06981}.

\bibitem[{Li et~al.(2023)Li, Patel, Vi{\'e}gas, Pfister, and
  Wattenberg}]{li2024inference}
Kenneth Li, Oam Patel, Fernanda Vi{\'e}gas, Hanspeter Pfister, and Martin
  Wattenberg. 2023.
\newblock \href {https://arxiv.org/abs/2306.03341} {Inference-time
  intervention: Eliciting truthful answers from a language model}.
\newblock In \emph{Advances in Neural Information Processing Systems
  (NeurIPS)}.

\bibitem[{Lieberum et~al.(2024)Lieberum, Rajamanoharan, Conmy, Smith, Sonnerat,
  Varma, Kram{\'a}r, Dragan, Shah, and Nanda}]{lieberum2024gemma}
Tom Lieberum, Senthooran Rajamanoharan, Arthur Conmy, Lewis Smith, Nicolas
  Sonnerat, Vikrant Varma, J{\'a}nos Kram{\'a}r, Anca Dragan, Rohin Shah, and
  Neel Nanda. 2024.
\newblock \href {https://arxiv.org/abs/2408.05147} {Gemma scope: Open sparse
  autoencoders everywhere all at once on {Gemma} 2}.
\newblock In \emph{BlackboxNLP Workshop}, pages 278--300.

\bibitem[{Lim et~al.(2025)Lim, Aji, and Cohn}]{lim2025language}
Zheng~Wei Lim, Alham~Fikri Aji, and Trevor Cohn. 2025.
\newblock \href {https://arxiv.org/abs/2505.13141} {Language-specific latent
  process hinders cross-lingual performance}.
\newblock \emph{arXiv preprint arXiv:2505.13141}.

\bibitem[{Machina and Mercer(2024)}]{machina2024anisotropy}
Anemily Machina and Robert Mercer. 2024.
\newblock \href {https://doi.org/10.18653/v1/2024.naacl-long.274} {Anisotropy
  is not inherent to transformers}.
\newblock In \emph{North American Chapter of the Association for Computational
  Linguistics (NAACL)}, pages 4892--4907.

\bibitem[{Marks and Tegmark(2024)}]{marks2024geometry}
Samuel Marks and Max Tegmark. 2024.
\newblock \href {https://arxiv.org/abs/2310.06824} {The geometry of truth:
  Emergent linear structure in large language model representations of
  true/false datasets}.
\newblock In \emph{Conference on Language Modeling (COLM)}.

\bibitem[{Martin and Mahoney(2019)}]{martin2019traditional}
Charles~H. Martin and Michael~W. Mahoney. 2019.
\newblock \href {https://arxiv.org/abs/1901.08276} {Traditional and
  heavy-tailed self regularization in neural network models}.
\newblock In \emph{International Conference on Machine Learning (ICML)}, pages
  4284--4293.

\bibitem[{Martin and Mahoney(2021)}]{martin2021implicit}
Charles~H. Martin and Michael~W. Mahoney. 2021.
\newblock \href {https://arxiv.org/abs/1810.01075} {Implicit
  self-regularization in deep neural networks: Evidence from random matrix
  theory and implications for learning}.
\newblock \emph{Journal of Machine Learning Research}, 22(165):1--73.

\bibitem[{Martin et~al.(2021)Martin, Peng, and Mahoney}]{martin2021predicting}
Charles~H. Martin, Tongsu Peng, and Michael~W. Mahoney. 2021.
\newblock \href {https://doi.org/10.1038/s41467-021-24025-8} {Predicting trends
  in the quality of state-of-the-art neural networks without access to training
  or testing data}.
\newblock \emph{Nature Communications}, 12:4122.

\bibitem[{Mikolov et~al.(2013)Mikolov, Chen, Corrado, and
  Dean}]{mikolov2013efficient}
Tomas Mikolov, Kai Chen, Greg Corrado, and Jeffrey Dean. 2013.
\newblock \href {https://arxiv.org/abs/1301.3781} {Efficient estimation of word
  representations in vector space}.
\newblock In \emph{ICLR Workshop}.

\bibitem[{Nanda et~al.(2023)Nanda, Lee, and Wattenberg}]{nanda2023emergent}
Neel Nanda, Andrew Lee, and Martin Wattenberg. 2023.
\newblock \href {https://doi.org/10.18653/v1/2023.blackboxnlp-1.2} {Emergent
  linear representations in world models of self-supervised sequence models}.
\newblock In \emph{BlackboxNLP Workshop}, pages 16--30.

\bibitem[{Nivre et~al.(2016)Nivre, de~Marneffe, Ginter, Goldberg, Haji{\v{c}},
  Manning, McDonald, Petrov, Pyysalo, Silveira, Tsarfaty, and
  Zeman}]{nivre2016universal}
Joakim Nivre, Marie-Catherine de~Marneffe, Filip Ginter, Yoav Goldberg, Jan
  Haji{\v{c}}, Christopher~D Manning, Ryan McDonald, Slav Petrov, Sampo
  Pyysalo, Natalia Silveira, Reut Tsarfaty, and Daniel Zeman. 2016.
\newblock \href {https://aclanthology.org/L16-1262/} {Universal dependencies
  v1: A multilingual treebank collection}.
\newblock In \emph{Proceedings of the Tenth International Conference on
  Language Resources and Evaluation (LREC)}, pages 1659--1666.

\bibitem[{Panickssery et~al.(2024)Panickssery, Gabrieli, Schulz, Tong,
  Hubinger, and Turner}]{panickssery2024steering}
Nina Panickssery, Nick Gabrieli, Julian Schulz, Meg Tong, Evan Hubinger, and
  Alexander~Matt Turner. 2024.
\newblock \href {https://doi.org/10.18653/v1/2024.acl-long.828} {Steering
  {Llama} 2 via contrastive activation addition}.
\newblock In \emph{Annual Meeting of the Association for Computational
  Linguistics (ACL)}, pages 15504--15522.

\bibitem[{Papyan et~al.(2020)Papyan, Han, and Donoho}]{papyan2020prevalence}
Vardan Papyan, X.Y. Han, and David~L. Donoho. 2020.
\newblock \href {https://doi.org/10.1073/pnas.2015509117} {Prevalence of neural
  collapse during the terminal phase of deep learning training}.
\newblock \emph{Proceedings of the National Academy of Sciences},
  117(40):24652--24663.

\bibitem[{Park et~al.(2025)Park, Choe, Jiang, and Veitch}]{park2025geometry}
Kiho Park, Yo~Joong Choe, Yibo Jiang, and Victor Veitch. 2025.
\newblock \href {https://arxiv.org/abs/2406.01506} {The geometry of categorical
  and hierarchical concepts in large language models}.
\newblock In \emph{International Conference on Learning Representations
  (ICLR)}.

\bibitem[{Park et~al.(2024)Park, Choe, and Veitch}]{park2024linear}
Kiho Park, Yo~Joong Choe, and Victor Veitch. 2024.
\newblock \href {https://arxiv.org/abs/2311.03658} {The linear representation
  hypothesis and the geometry of large language models}.
\newblock In \emph{International Conference on Machine Learning (ICML)}, volume
  235 of \emph{PMLR}, pages 39643--39666.

\bibitem[{Puccetti et~al.(2022)Puccetti, Rogers, Drozd, and
  Dell'Orletta}]{puccetti2022outlier}
Giovanni Puccetti, Anna Rogers, Aleksandr Drozd, and Felice Dell'Orletta. 2022.
\newblock \href {https://doi.org/10.18653/v1/2022.findings-emnlp.93} {Outlier
  dimensions that disrupt transformers are driven by frequency}.
\newblock In \emph{Findings of EMNLP}, pages 1286--1304.

\bibitem[{Scherlis et~al.(2022)Scherlis, Sachan, Jermyn, Benton, and
  Shlegeris}]{scherlis2022polysemanticity}
Adam Scherlis, Kshitij Sachan, Adam~S Jermyn, Joe Benton, and Buck Shlegeris.
  2022.
\newblock \href {https://arxiv.org/abs/2210.01892} {Polysemanticity and
  capacity in neural networks}.
\newblock \emph{arXiv preprint arXiv:2210.01892}.

\bibitem[{Sharma et~al.(2023)Sharma, Ash, and Misra}]{sharma2023truth}
Pratyusha Sharma, Jordan~T. Ash, and Dipendra Misra. 2023.
\newblock \href {https://arxiv.org/abs/2312.13558} {The truth is in there:
  Improving reasoning in language models with layer-selective rank reduction}.
\newblock \emph{arXiv preprint arXiv:2312.13558}.

\bibitem[{Staats et~al.(2024)Staats, Thamm, and Rosenow}]{staats2024small}
Max Staats, Matthias Thamm, and Bernd Rosenow. 2024.
\newblock \href {https://arxiv.org/abs/2410.17770} {Small singular values
  matter: A random matrix analysis of transformer models}.
\newblock \emph{arXiv preprint arXiv:2410.17770}.

\bibitem[{Su et~al.(2021)Su, Cao, Liu, and Ou}]{su2021whitening}
Jianlin Su, Jiarun Cao, Weijie Liu, and Yangyiwen Ou. 2021.
\newblock \href {https://arxiv.org/abs/2103.15316} {Whitening sentence
  representations for better semantics and faster retrieval}.
\newblock \emph{arXiv preprint arXiv:2103.15316}.

\bibitem[{Subramani et~al.(2022)Subramani, Suresh, and
  Peters}]{subramani2022extracting}
Nishant Subramani, Nivedita Suresh, and Matthew~E. Peters. 2022.
\newblock \href {https://doi.org/10.18653/v1/2022.findings-acl.48} {Extracting
  latent steering vectors from pretrained language models}.
\newblock In \emph{Findings of ACL}, pages 566--581.

\bibitem[{Templeton et~al.(2024)Templeton, Conerly, Marcus, Lindsey, Bricken,
  Chen et~al.}]{templeton2024scaling}
Adly Templeton, Tom Conerly, Jonathan Marcus, Jack Lindsey, Trenton Bricken,
  Brian Chen, and 1 others. 2024.
\newblock \href
  {https://transformer-circuits.pub/2024/scaling-monosemanticity/index.html}
  {Scaling monosemanticity: Extracting interpretable features from {Claude} 3
  {Sonnet}}.
\newblock \emph{Transformer Circuits Thread}.

\bibitem[{Tigges et~al.(2024)Tigges, Hollinsworth, Geiger, and
  Nanda}]{tigges2024linear}
Curt Tigges, Oskar~John Hollinsworth, Atticus Geiger, and Neel Nanda. 2024.
\newblock \href {https://arxiv.org/abs/2310.15154} {Linear representations of
  sentiment in large language models}.
\newblock In \emph{BlackboxNLP Workshop}, pages 58--87.

\bibitem[{Timkey and van Schijndel(2021)}]{timkey2021all}
William Timkey and Marten van Schijndel. 2021.
\newblock \href {https://doi.org/10.18653/v1/2021.emnlp-main.372} {All bark and
  no bite: Rogue dimensions in transformer language models obscure
  representational quality}.
\newblock In \emph{Conference on Empirical Methods in Natural Language
  Processing (EMNLP)}, pages 4527--4546.

\bibitem[{Turner et~al.(2023)Turner, Thiergart, Udell, Leech, Mini, and
  MacDiarmid}]{turner2024activation}
Alexander~Matt Turner, Lisa Thiergart, David Udell, Gavin Leech, Ulisse Mini,
  and Monte MacDiarmid. 2023.
\newblock \href {https://arxiv.org/abs/2308.10248} {Activation addition:
  Steering language models without optimization}.
\newblock \emph{arXiv preprint arXiv:2308.10248}.

\bibitem[{Wendler et~al.(2024)Wendler, Veselovsky, Monea, and
  West}]{wendler2024llamas}
Chris Wendler, Veniamin Veselovsky, Giovanni Monea, and Robert West. 2024.
\newblock \href {https://doi.org/10.18653/v1/2024.acl-long.820} {Do llamas work
  in english? on the latent language of multilingual transformers}.
\newblock In \emph{Annual Meeting of the Association for Computational
  Linguistics (ACL)}, pages 15366--15394.

\bibitem[{Wu and Papyan(2024)}]{wu2024linguistic}
Robert Wu and Vardan Papyan. 2024.
\newblock \href {https://arxiv.org/abs/2405.17767} {Linguistic collapse: Neural
  collapse in (large) language models}.
\newblock \emph{arXiv preprint arXiv:2405.17767}.

\bibitem[{Zou et~al.(2023)Zou, Phan, Chen, Campbell, Guo, Ren
  et~al.}]{zou2023representation}
Andy Zou, Long Phan, Sarah Chen, James Campbell, Phillip Guo, Richard Ren, and
  1 others. 2023.
\newblock \href {https://arxiv.org/abs/2310.01405} {Representation engineering:
  A top-down approach to {AI} transparency}.
\newblock \emph{arXiv preprint arXiv:2310.01405}.

\end{thebibliography}

\appendix

\section{Hyperparameters and Compute}
\label{app:hyperparams}

\paragraph{Spectral analysis.} Regularization $\tau=0.1$ unless stated; eigendecomposition via dense SVD on $\Sigma$. Random baselines use 1{,}000 unit vectors sampled uniformly from $S^{d-1}$.

\paragraph{WCA.} 5 random seeds, 17 models $\times$ 4 language pairs (en-fr, es-de, fr-de, fr-es; three of the four non-English), $\tau=0.1$. Procrustes solved on whitened space via SVD.

\paragraph{Linear probes.} L2-regularized logistic regression ($C=1.0$, max\_iter $=2000$), 5-fold stratified CV, per-model standardization disabled (uncentered).

\paragraph{SAE.} Gemma Scope and Llama Scope public releases (no retraining). Top-$k\in\{1,3,5\}$ differentially activating features; decoder-column unit vectors.

\paragraph{Split-injection.} Layers 50/62/75/87\% of depth; $\alpha\in\{1,5,10,20,50,100\}$ swept (full sweep in \Cref{tab:split-injection-full}); perplexity on 10 held-out Wikipedia passages.

\paragraph{POS probing.} UD English-EWT dev, first 500 sentences (7{,}621 syntactic words; 6{,}585 after excluding \textsc{punct}/\textsc{sym}/\textsc{x} and tags with $<30$ samples, leaving 13 classes); UD Chinese-GSD, all 500 dev sentences (12{,}665 words; 10{,}813 after the same filtering); $k\in\{5\%,10\%,20\%\}$; 5-fold stratified CV plus 10{,}000-resample token-level bootstrap.

\paragraph{Compute.} Experiments were run on a mix of NVIDIA RTX 6000 Ada and RTX A6000 GPUs. The full pipeline (17 models, 27 concepts, three extraction methods, WCA randomization, split-injection sweeps, POS probing) consumed approximately 320 GPU-hours; the dominant cost is split-injection on Llama-3.1-8B and Qwen2.5-14B ($\sim$110 GPU-hours combined).

\section{Full Model Details}
\label{app:models}

\begin{table*}[t]
\centering
\caption{Complete model specifications for all 18 models tested (17 in the core spectral analysis, plus Llama-3.1-8B used for SAE and functional tests).}
\label{tab:full-models}
\small
\begin{tabular}{llcccc}
\toprule
\textbf{Family} & \textbf{Model} & \textbf{Params} & \textbf{$d$} & \textbf{Vocab} & \textbf{Cond.\ \#} \\
\midrule
\multirow{2}{*}{Llama 3.2} & Llama-3.2-1B & 1B & 2048 & 128K & $1.4 \times 10^4$ \\
  & Llama-3.2-3B & 3B & 3072 & 128K & $2.5 \times 10^4$ \\
\midrule
Llama 3.1 & Llama-3.1-8B & 8B & 4096 & 128K & $2.7 \times 10^4$ \\
\midrule
\multirow{5}{*}{Qwen 2.5} & Qwen2.5-0.5B & 0.5B & 896 & 152K & $4.2 \times 10^3$ \\
  & Qwen2.5-1.5B & 1.5B & 1536 & 152K & $5.8 \times 10^3$ \\
  & Qwen2.5-3B & 3B & 2048 & 152K & $5.7 \times 10^3$ \\
  & Qwen2.5-7B & 7B & 3584 & 152K & $2.9 \times 10^4$ \\
  & Qwen2.5-14B & 14B & 5120 & 152K & $1.6 \times 10^4$ \\
\midrule
\multirow{4}{*}{Qwen 3} & Qwen3-0.6B & 0.6B & 1024 & 152K & $2.6 \times 10^3$ \\
  & Qwen3-1.7B & 1.7B & 2048 & 152K & $4.3 \times 10^3$ \\
  & Qwen3-4B & 4B & 2560 & 152K & $1.0 \times 10^4$ \\
  & Qwen3-8B & 8B & 4096 & 152K & $1.6 \times 10^4$ \\
\midrule
\multirow{2}{*}{Gemma 2} & Gemma-2-2b & 2B & 2304 & 256K & $1.6 \times 10^5$ \\
  & Gemma-2-9b & 9B & 3584 & 256K & $1.6 \times 10^5$ \\
\midrule
Mistral & Mistral-7B-v0.3 & 7B & 4096 & 32K & $2.8 \times 10^5$ \\
\midrule
SmolLM & SmolLM2-1.7B & 1.7B & 2048 & 49K & $2.7 \times 10^4$ \\
\midrule
\multirow{2}{*}{MoE} & OLMoE-1B-7B & 1B/7B & 2048 & 50K & $7.7 \times 10^3$ \\
  & JetMoE-8B & 8B & 2048 & 32K & $3.7 \times 10^3$ \\
\bottomrule
\end{tabular}
\end{table*}

\Cref{tab:full-models} lists every model with its parameter count, hidden dimension, vocabulary size and condition number. All models are decoder-only transformers with pre-normalization (RMSNorm for Gemma, Llama, Qwen, and Mistral; LayerNorm for SmolLM2 and the MoE models). Concept vectors and covariance matrices are computed from the unembedding matrix $W_U$ accessed via each model's LM head; for tied-weight models (Qwen, SmolLM2) $W_U$ is the shared embedding matrix. All experiments use base (non-instruct) variants.

\section{Full Spectral Metrics (All 17 Models)}
\label{app:full-tables}

\begin{table*}[t]
\centering
\caption{Spectral Center of Mass and Gini deviation for all 17 models. Mean across models is reported on the last row.}
\label{tab:full-scm}
\small
\begin{tabular}{lcccccc}
\toprule
\textbf{Model} & \textbf{$d$} & \textbf{Cond.\ \#} & \textbf{Concept \scm} & \textbf{Random \scm} & \textbf{Gap} & \textbf{Gini} \\
\midrule
Qwen2.5-0.5B   & 896  & $4.2\text{K}$  & 0.997 & 0.711 & $+$0.286 & $-$0.356 \\
Qwen2.5-1.5B   & 1536 & $5.8\text{K}$  & 0.986 & 0.729 & $+$0.257 & $-$0.282 \\
Qwen2.5-3B     & 2048 & $5.7\text{K}$  & 0.973 & 0.735 & $+$0.238 & $-$0.272 \\
Qwen2.5-7B     & 3584 & $29\text{K}$   & 0.999 & 0.757 & $+$0.242 & $-$0.396 \\
Qwen2.5-14B    & 5120 & $16\text{K}$   & 0.996 & 0.790 & $+$0.205 & $-$0.346 \\
Qwen3-0.6B     & 1024 & $2.6\text{K}$  & 0.938 & 0.723 & $+$0.215 & $-$0.216 \\
Qwen3-1.7B     & 2048 & $4.3\text{K}$  & 0.997 & 0.738 & $+$0.259 & $-$0.248 \\
Qwen3-4B       & 2560 & $10\text{K}$   & 0.994 & 0.750 & $+$0.243 & $-$0.312 \\
Qwen3-8B       & 4096 & $16\text{K}$   & 1.000 & 0.783 & $+$0.217 & $-$0.359 \\
Llama-3.2-1B   & 2048 & $14\text{K}$   & 0.738 & 0.748 & $-$0.010 & $-$0.140 \\
Llama-3.2-3B   & 3072 & $25\text{K}$   & 0.771 & 0.771 & $+$0.001 & $-$0.159 \\
Gemma-2-2b     & 2304 & $160\text{K}$  & 0.964 & 0.796 & $+$0.168 & $-$0.389 \\
Gemma-2-9b     & 3584 & $160\text{K}$  & 0.927 & 0.849 & $+$0.078 & $-$0.364 \\
Mistral-7B     & 4096 & $284\text{K}$  & 0.908 & 0.792 & $+$0.116 & $-$0.199 \\
SmolLM2-1.7B   & 2048 & $27\text{K}$   & 1.000 & 0.735 & $+$0.265 & $-$0.436 \\
OLMoE-1B-7B    & 2048 & $7.7\text{K}$  & 0.620 & 0.736 & $-$0.116 & $-$0.076 \\
JetMoE-8B      & 2048 & $3.7\text{K}$  & 0.945 & 0.741 & $+$0.205 & $-$0.242 \\
\midrule
\textbf{Mean}  &      &                & 0.926 & 0.758 & $+$0.169 & $-$0.282 \\
\bottomrule
\end{tabular}
\end{table*}

\Cref{tab:full-scm} gives the per-model figures. All 17 models show negative Gini, supporting broad anti-concentration. The Llama~3.2 family shows the smallest \scm gaps ($-0.010$ to $+0.001$); SmolLM2 has the strongest Gini ($-0.436$); OLMoE has the largest negative \scm gap ($-0.116$; Llama-3.2-1B is also slightly negative, $-0.010$) while still anti-concentrating (Gini $=-0.076$). The correlation between random and concept \scm is negligible ($r=0.019$, $p=0.94$).

\section{Expanded Semantic Concept Set}
\label{app:expanded-semantic}

To test whether anti-concentration depends on the original narrow semantic subset (7 of 27 concepts), we added 15 further semantic categories from BATS~\citep{gladkova2016analogy}---encyclopedic (country--language, name--nationality, name--occupation, animal--young, animal--sound, animal--shelter) and lexicographic (hypernyms$\times$2, hyponyms, meronyms$\times$2, synonyms$\times$2, antonyms$\times$2)---selected by a uniform rule (first single-token target per pair), more than tripling the semantic set from 7 to 22. We recomputed \dom concept directions and their Gini deviations against the same en-fr covariance eigenbasis for all 17 core models.

\Cref{tab:expanded-semantic} reports per-model results. Taking the model as the unit of analysis, the expanded set has mean Gini $-0.278$, statistically identical to the original semantic subset ($-0.278$; Wilcoxon signed-rank $p=0.71$), and all 17 models are negative (one-sample $t=-12.3$, $p=1.5\times10^{-9}$). A few models retain $n<15$ expanded concepts because some BATS words are multi-token and yield zero-norm vectors, exactly as for the core set on Gemma (\Cref{app:hole4-details}).

\paragraph{Instruction-tuned models.}
Anti-concentration persists under instruction tuning: on Qwen2.5-3B-Instruct the expanded-set Gini is $-0.281$ (vs.\ $-0.280$ for the base model) and on Llama-3.2-3B-Instruct it is $-0.182$ (vs.\ $-0.167$). gemma-2-2b-it is omitted because its chat template renders most single-token pairs multi-token, leaving too few concepts to estimate reliably.

\paragraph{Centered covariance.}
Recomputing the expanded-set Gini against the \emph{centered} covariance (\Cref{eq:cov} with the mean subtracted, matching \citealp{park2024linear}) gives mean $-0.288$ (vs.\ $-0.278$ uncentered; $t=-11.8$, $p=2.6\times10^{-9}$), so the effect is, if anything, slightly stronger under centering.

\begin{table}[t]
\centering
\small
\caption{Expanded semantic set (15 added BATS categories) vs.\ the original 7 semantic concepts, per model. $n$ is the number of expanded concepts with non-zero-norm vectors. The Exp.\ Gini column averages the 15 \emph{added} categories only; the 22-concept set quoted in \Cref{sec:anti-concentration} is their union with the original 7. All expanded means are negative (anti-concentrated).}
\label{tab:expanded-semantic}
\begin{tabular}{lccc}
\toprule
\textbf{Model} & \textbf{$n$} & \textbf{Exp.\ Gini} & \textbf{Orig.\ Gini} \\
\midrule
Qwen2.5-0.5B    & 15 & $-0.345$ & $-0.347$ \\
Qwen2.5-1.5B    & 15 & $-0.288$ & $-0.280$ \\
Qwen2.5-3B      & 15 & $-0.280$ & $-0.268$ \\
Qwen2.5-7B      & 15 & $-0.392$ & $-0.404$ \\
Qwen2.5-14B     & 15 & $-0.335$ & $-0.347$ \\
Qwen3-0.6B      & 15 & $-0.228$ & $-0.249$ \\
Qwen3-1.7B      & 15 & $-0.224$ & $-0.268$ \\
Qwen3-4B        & 15 & $-0.288$ & $-0.295$ \\
Qwen3-8B        & 15 & $-0.353$ & $-0.339$ \\
Llama-3.2-1B    & 15 & $-0.142$ & $-0.139$ \\
Llama-3.2-3B    & 15 & $-0.167$ & $-0.169$ \\
gemma-2-2b      & 13 & $-0.330$ & $-0.336$ \\
gemma-2-9b      & 13 & $-0.314$ & $-0.311$ \\
Mistral-7B-v0.3 & 15 & $-0.212$ & $-0.174$ \\
SmolLM2-1.7B    & 15 & $-0.455$ & $-0.437$ \\
OLMoE-1B-7B     & 15 & $-0.086$ & $-0.070$ \\
JetMoE-8B       & 11 & $-0.291$ & $-0.285$ \\
\midrule
\textbf{Mean}   &    & $\mathbf{-0.278}$ & $\mathbf{-0.278}$ \\
\bottomrule
\end{tabular}
\end{table}

\section{Robustness and Layer-wise Controls}
\label{app:robustness-controls}

We stress-test the anti-concentration result under covariance variants, random-direction nulls, and layer-wise trajectories. These controls use the same concept set as the core spectral analysis.

\begin{figure*}[t]
  \centering
  \includegraphics[width=\textwidth]{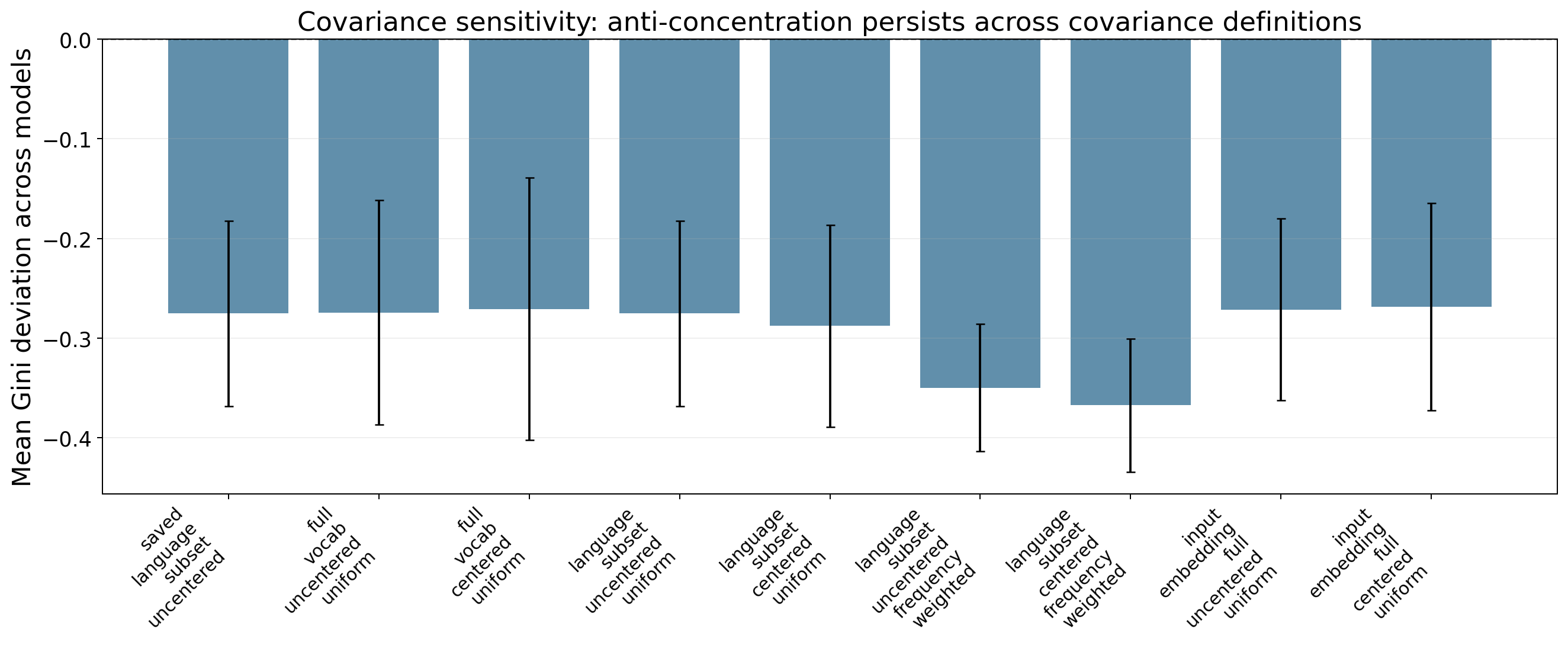}
  \caption{Covariance robustness. Mean Gini remains negative across covariance construction choices, including centered and frequency-weighted variants and input-embedding spectra. Error bars show variation across models.}
  \label{fig:v2-covariance-robustness}
\end{figure*}

\paragraph{Covariance variants.}
The effect persists across full-vocabulary, language-subset, centered, uncentered, frequency-weighted, and input-embedding spectra (\Cref{fig:v2-covariance-robustness}). The saved language-subset covariance yields mean Gini $-0.275$ over 17 models; frequency weighting strengthens the effect (mean Gini $-0.350$ uncentered, $-0.367$ centered), arguing against the result being an artifact of uniformly weighting rare vocabulary items.

\begin{figure*}[t]
  \centering
  \begin{subfigure}[b]{0.49\textwidth}
    \includegraphics[width=\textwidth]{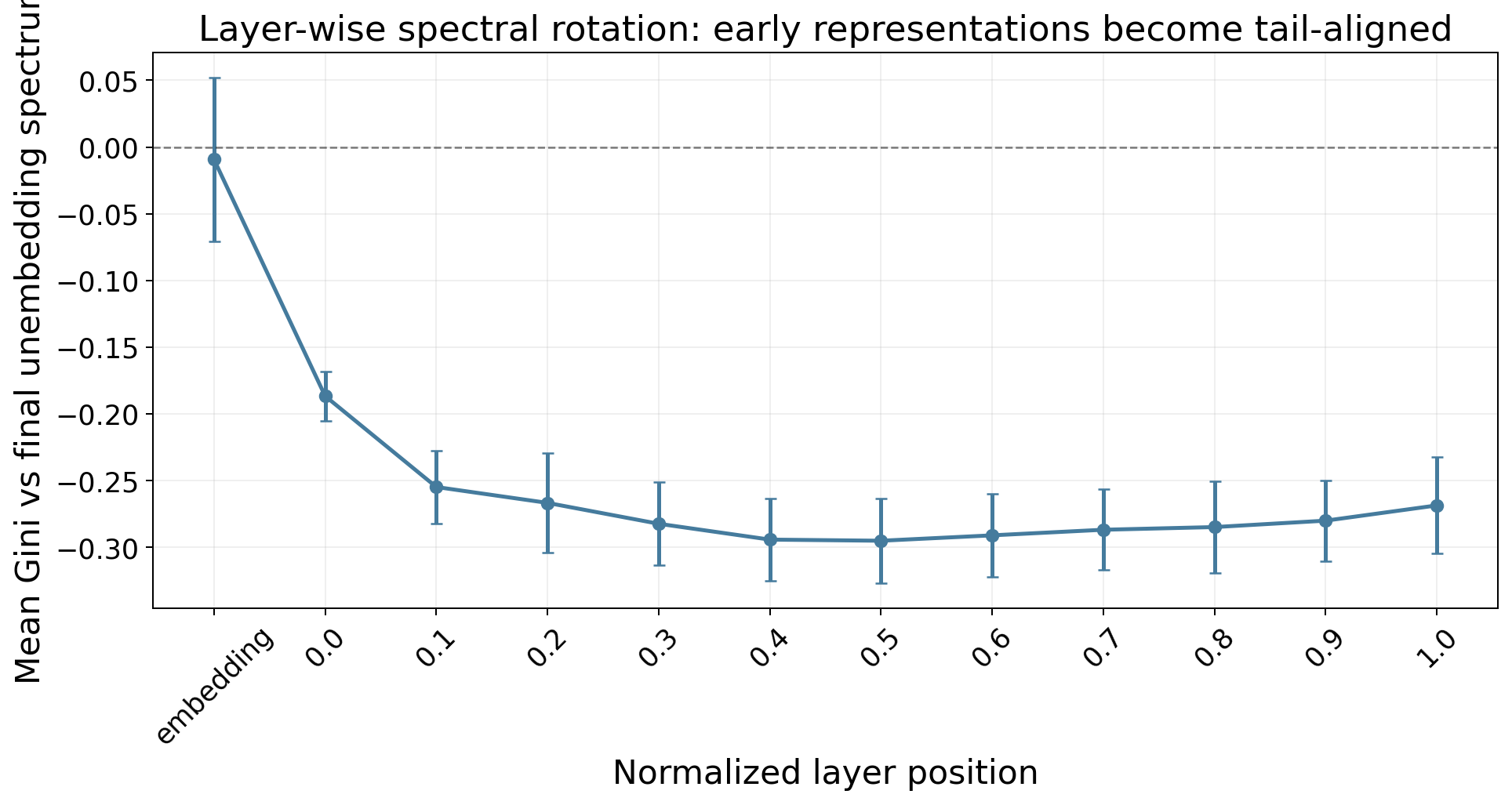}
    \caption{Aggregate layer-wise trajectory}
  \end{subfigure}\hfill
  \begin{subfigure}[b]{0.49\textwidth}
    \includegraphics[width=\textwidth]{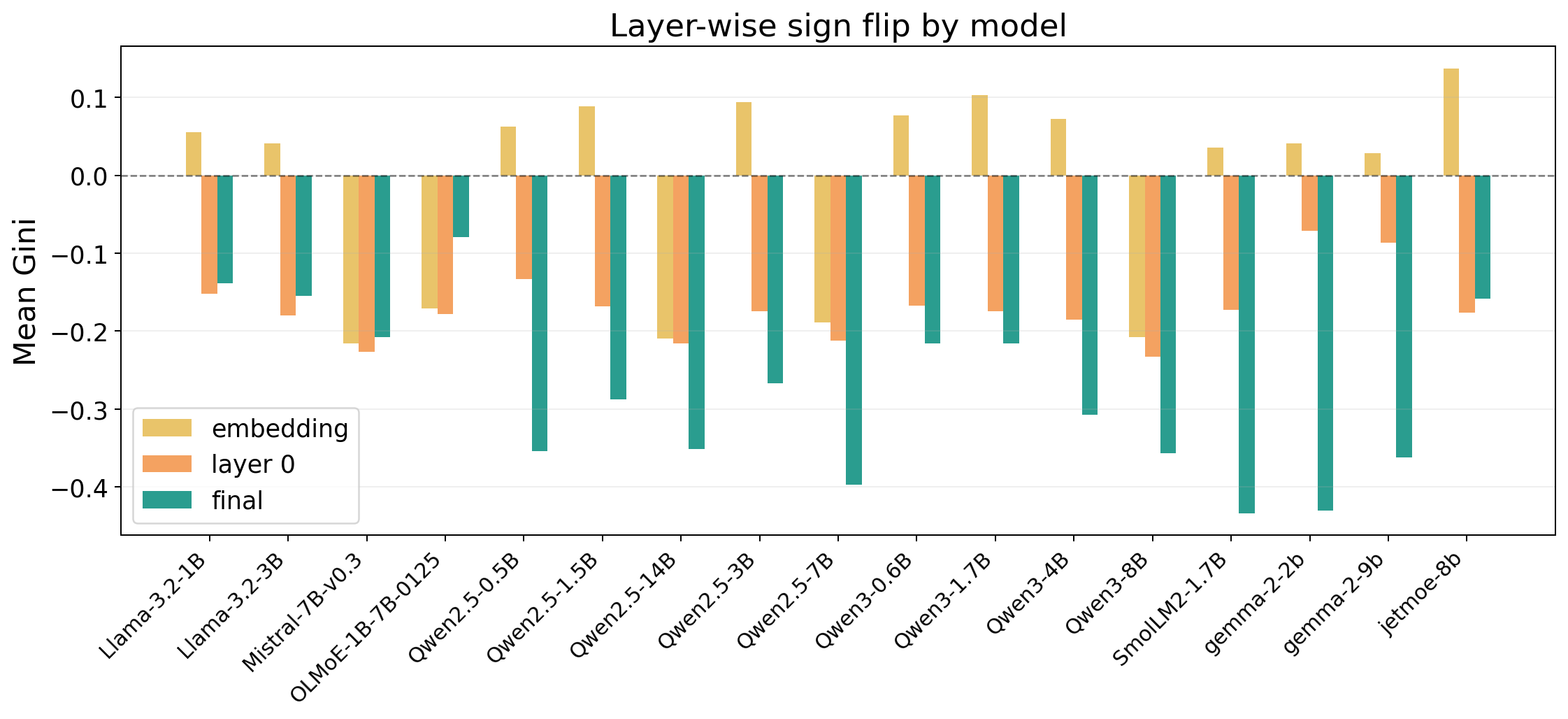}
    \caption{Embedding, layer~0, and final layer by model}
  \end{subfigure}
  \caption{Layer-wise spectral rotation. Concept directions move from near-neutral or mixed spectral placement at the embedding layer to stronger low-variance alignment after early transformer processing and by the final layer.}
  \label{fig:v2-layerwise}
\end{figure*}

\paragraph{Layer-wise trajectory.}
Concept directions are near-neutral at the embedding layer and become tail-aligned after contextualization (\Cref{fig:v2-layerwise}). Across models, mean Gini shifts from $-0.009$ at the embedding layer to $-0.171$ at layer~0 and $-0.277$ at the final layer; the trough is at mid-depth, $-0.295$ at 50\% of depth. This supports the interpretation that the forward pass rotates concept information from the vocabulary geometry into the low-variance activation subspace.

\begin{figure*}[t]
  \centering
  \includegraphics[width=\textwidth]{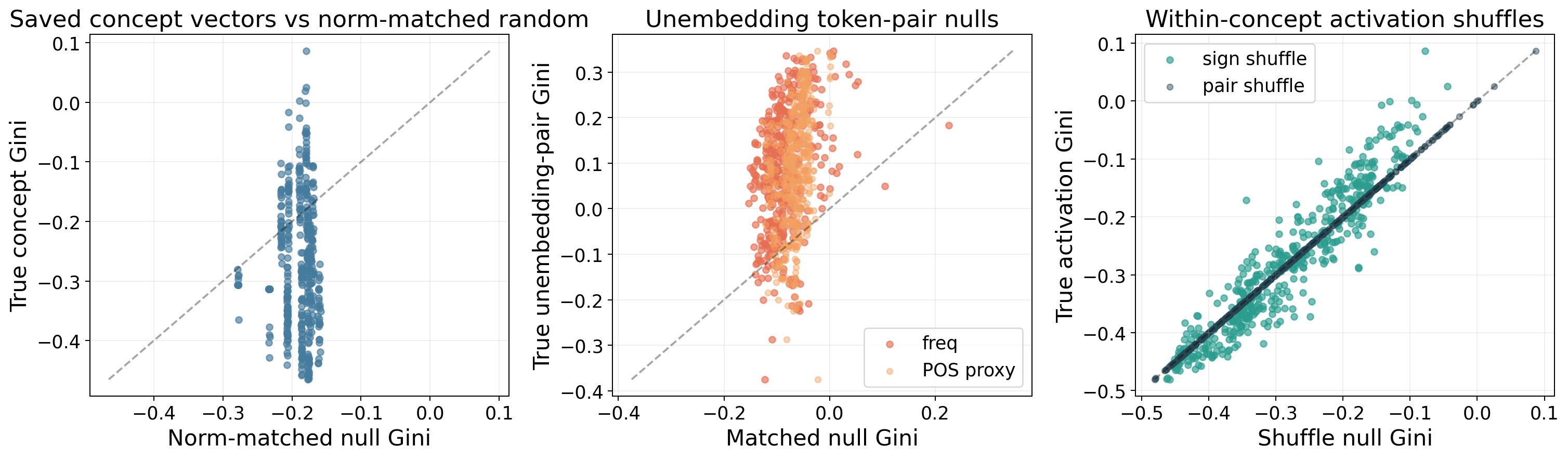}
  \caption{Norm-matched random-direction nulls. Most models place true concept directions further into the spectral tail than matched random directions; exceptions coincide with the weaker anti-concentration cases in the full spectral table.}
  \label{fig:v2-strong-nulls}
\end{figure*}

\paragraph{Random-direction nulls.}
Random directions evaluated in an anisotropic spectrum can themselves show negative Gini, so the relevant test is whether learned concept directions are \emph{more} tail-aligned than matched random directions. Norm-matched random-direction controls show a mean true-minus-null Gini gap of $-0.082$ across model summaries, with stronger tail alignment in 13/17 models (\Cref{fig:v2-strong-nulls}). Llama~3.2, Mistral, and OLMoE are weaker under this null, matching the smaller \scm gaps noted in \Cref{app:full-tables}.

\section{SAE Feature Collapse Analysis}
\label{app:sae-collapse}

\begin{figure*}[t]
  \centering
  \includegraphics[width=\textwidth]{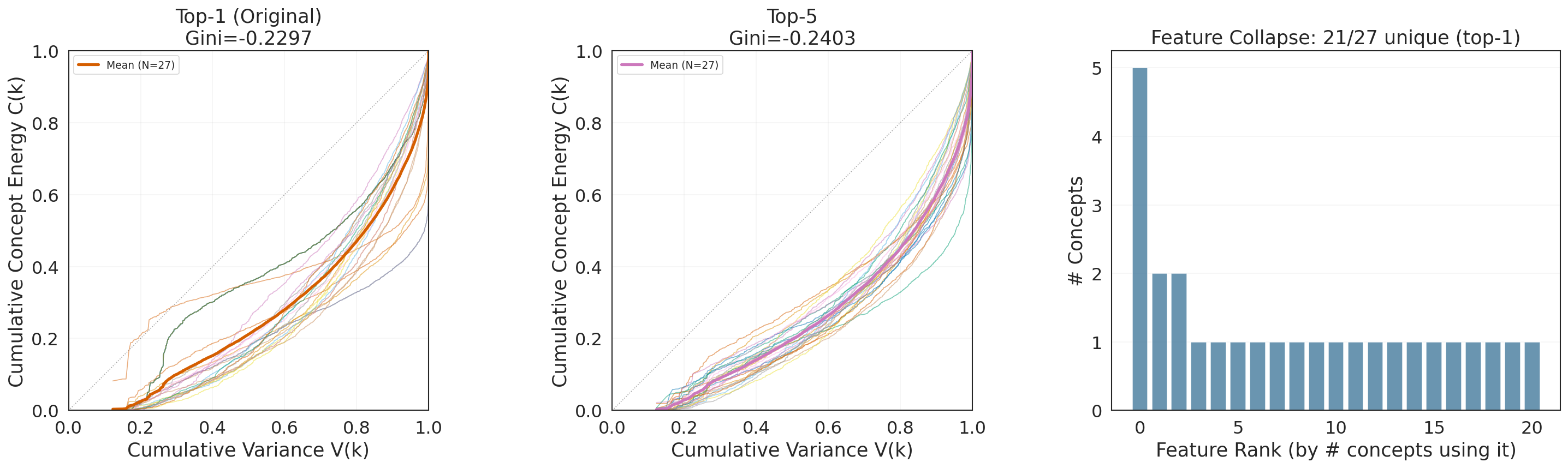}
  \caption{SAE feature collapse on gemma-2-2b. The most-collapsed feature (\#11527, 5/27 concepts) is the weakest; removing collapsed concepts strengthens the aggregate Gini.}
  \label{fig:sae-collapse}
\end{figure*}

On gemma-2-2b with top-1 selection, the 27 concepts mapped onto 21 distinct SAE features (77.8\%). Three features appeared more than once: \#11527 (5 concepts), \#38 (2), \#1643 (2). Feature \#11527 had the weakest Gini ($-0.147$). Two \emph{distinct} controls address collapse, and they happen to land on the same rounded value. First, removing all 9 concepts involved in any feature collapse ($5+2+2$, leaving an effective $N=18$) \emph{strengthened} the top-1 mean Gini from $-0.230$ to $-0.240$ ($t=-33.3$, $p=6.3\times 10^{-17}$). Second, and separately, top-5 selection (84/135 feature slots unique) gives an aggregate Gini of $-0.240$ ($p=1.2\times 10^{-27}$), stronger than top-1. On gemma-2-9b with top-5, all 27 concepts mapped to distinct feature sets with zero collapse. \Cref{fig:sae-collapse} plots the per-feature breakdown.

\section{Artifact-Check Per-Concept Plots}
\label{app:artifact-check}

\begin{figure*}[t]
  \centering
  \begin{subfigure}[b]{\textwidth}
    \centering
    \includegraphics[width=\textwidth]{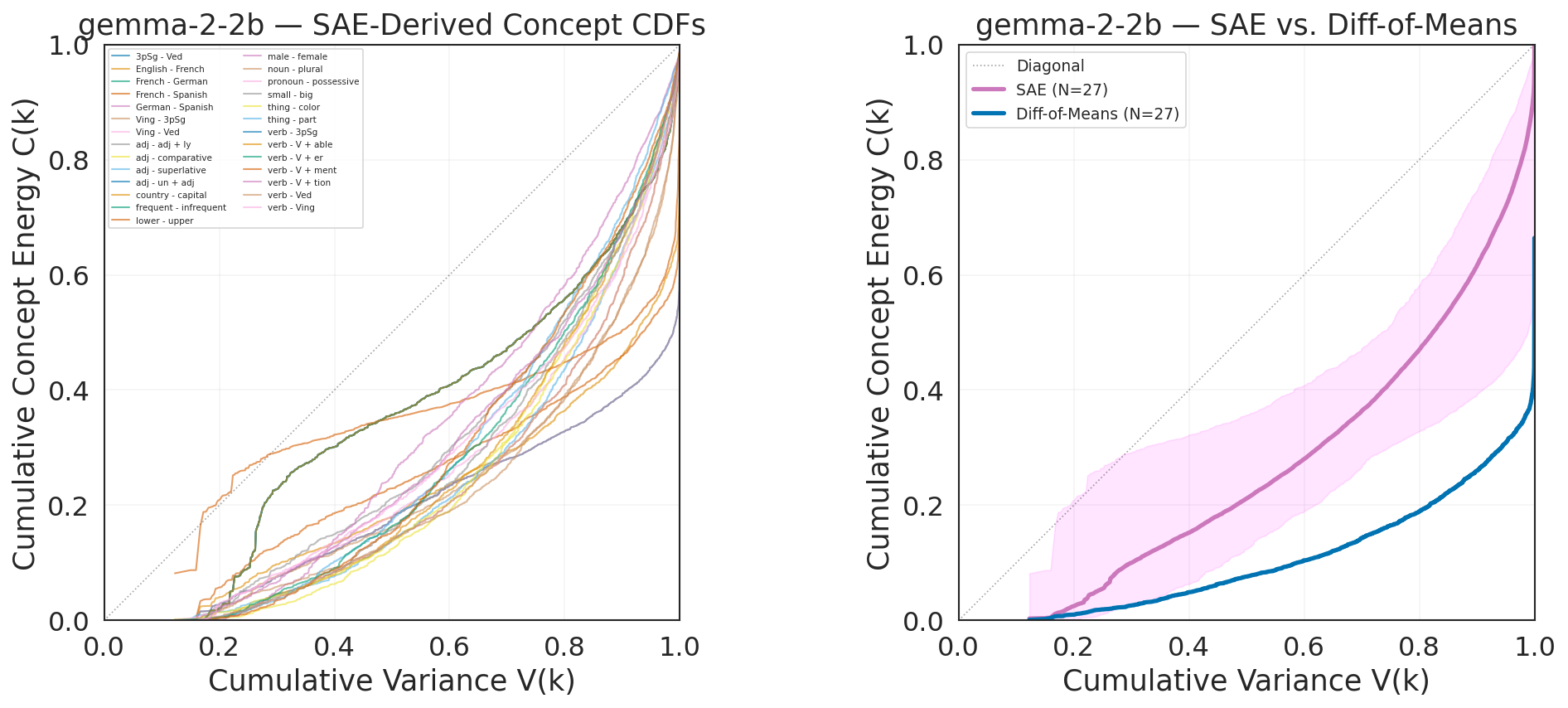}
    \caption{SAE features (gemma-2-2b)}
  \end{subfigure}
  \par\medskip
  \begin{subfigure}[b]{\textwidth}
    \centering
    \includegraphics[width=\textwidth]{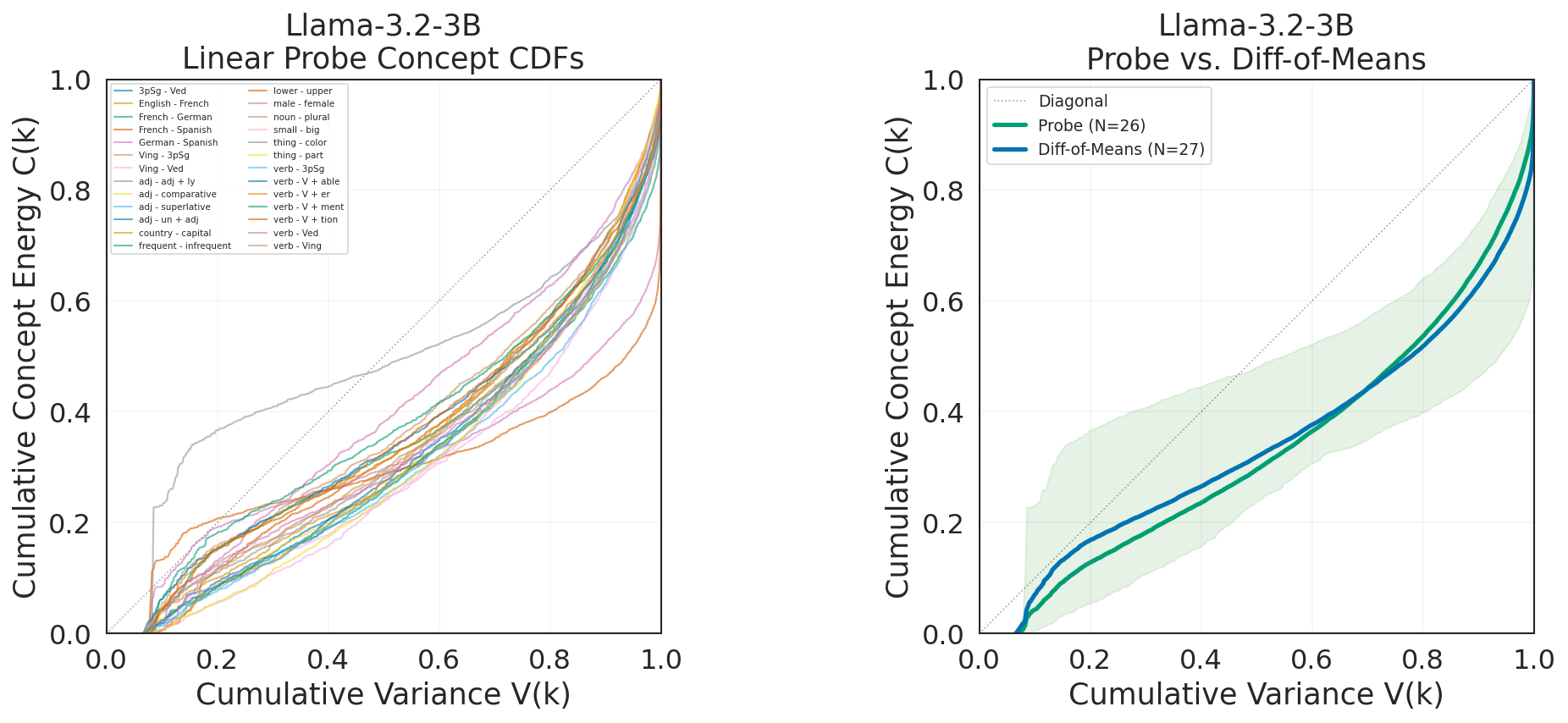}
    \caption{Linear probes (Llama-3.2-3B)}
  \end{subfigure}
  \caption{Per-concept spectral energy CDFs. Nearly all concept curves (colored) fall below the uniform diagonal (dotted) under both extraction methods; the right-hand panel of each pair shows the method mean with its min--max envelope across concepts, alongside the \dom curve for comparison.}
  \label{fig:artifact-check}
\end{figure*}

\begin{figure*}[t]
  \centering
  \begin{subfigure}[b]{\textwidth}
    \centering
    \includegraphics[width=\textwidth]{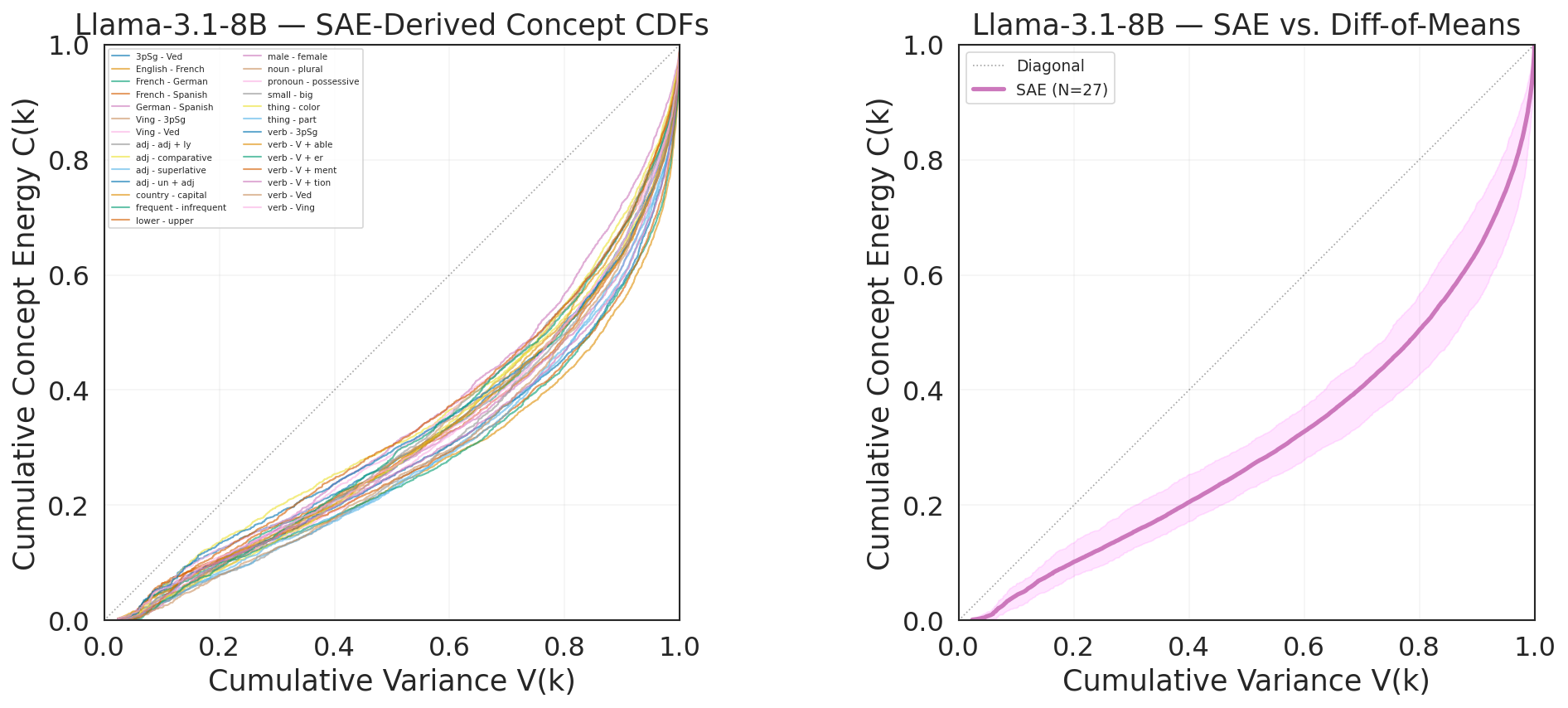}
    \caption{SAE features (Llama-3.1-8B)}
  \end{subfigure}
  \par\medskip
  \begin{subfigure}[b]{\textwidth}
    \centering
    \includegraphics[width=\textwidth]{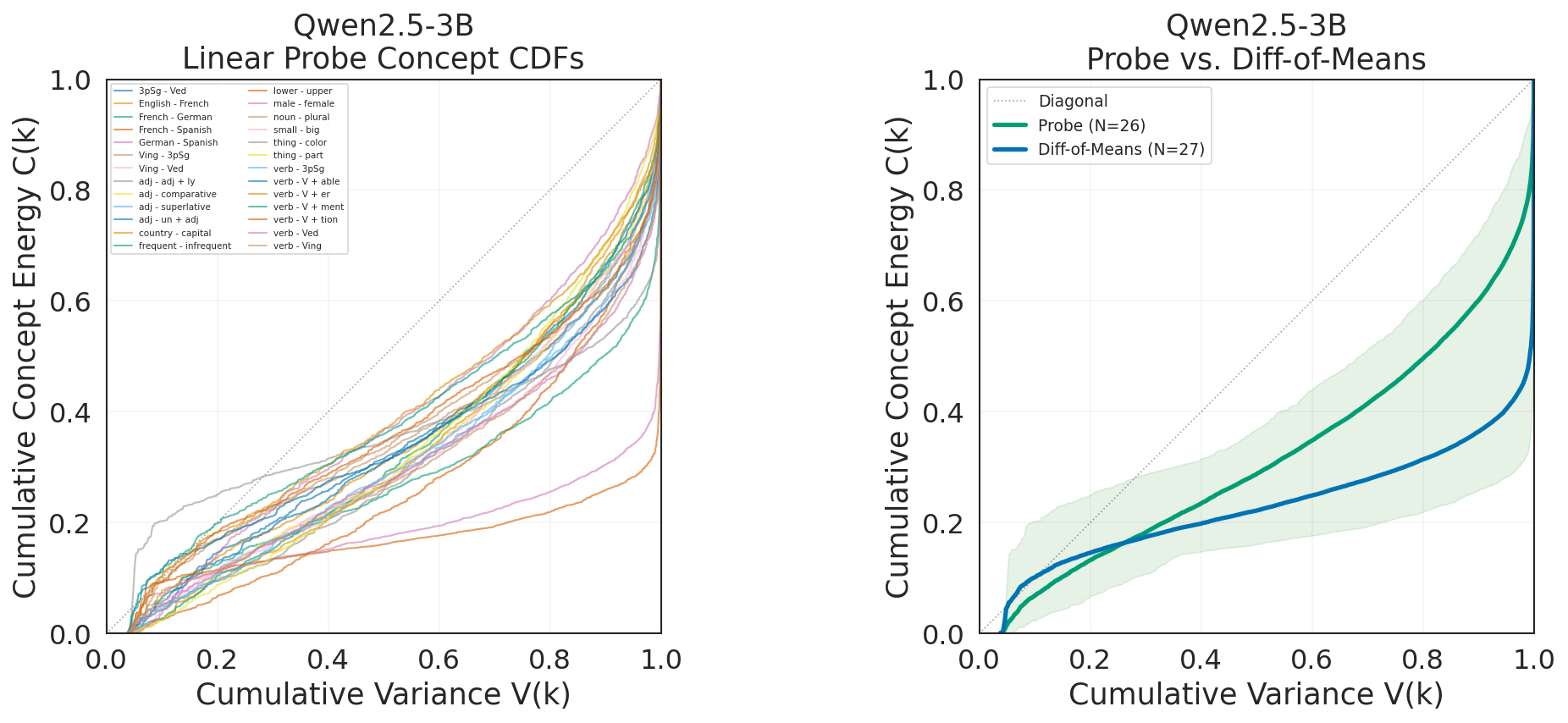}
    \caption{Linear probes (Qwen2.5-3B)}
  \end{subfigure}
  \caption{Cross-architecture replication. Top: SAE features on Llama-3.1-8B ($p=3.6\times 10^{-27}$). Bottom: linear probes on Qwen2.5-3B (Gini $=-0.185$, $p<10^{-15}$).}
  \label{fig:artifact-check-cross}
\end{figure*}

\begin{figure*}[t]
  \centering
  \includegraphics[width=0.85\textwidth]{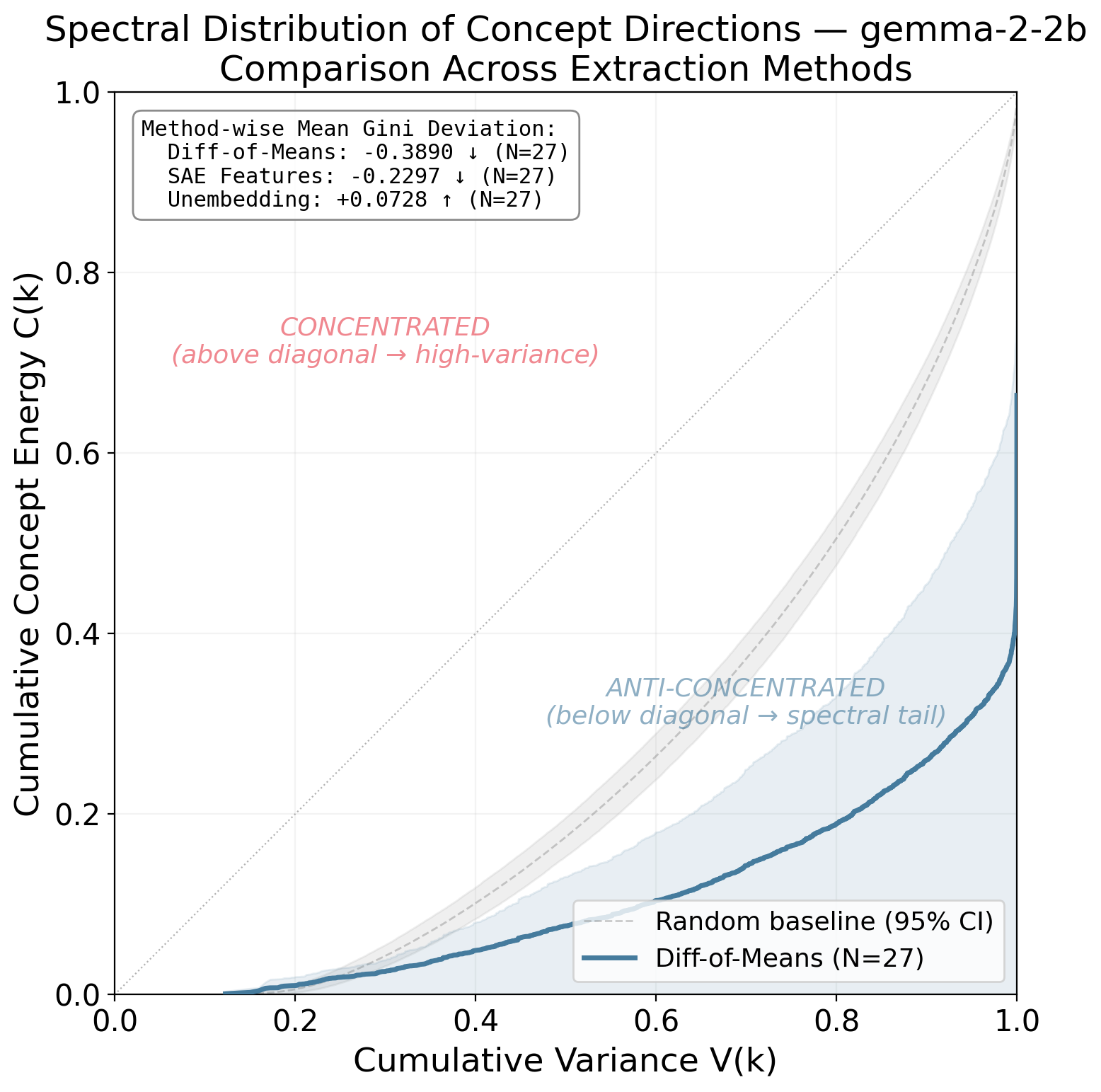}
  \caption{Cross-method spectral energy CDFs on gemma-2-2b: \dom, SAE features, and unembedding-row contrasts. Activation methods fall below the diagonal; unembedding rows flip sign---the dual geometry.}
  \label{fig:hole6_overlay_gemma-2-2b}
\end{figure*}

This appendix shows per-concept spectral energy CDFs for the non-subtractive extraction methods used in \Cref{sec:artifact-check}, as a visual check that anti-concentration is not driven by a few outlier concepts or by \dom subtraction. \Cref{fig:artifact-check} pairs SAE features on gemma-2-2b with linear probes on Llama-3.2-3B; \Cref{fig:artifact-check-cross} repeats both on a second architecture each (SAE on Llama-3.1-8B, probes on Qwen2.5-3B); and \Cref{fig:hole6_overlay_gemma-2-2b} overlays all three activation-space methods against the unembedding-row contrast on a single model, where the sign flip of the dual geometry is visible directly.

\section{Dual Geometry Details}
\label{app:dual-geometry}

This appendix reports the unembedding-row null used to support the dual-geometry claim in \Cref{sec:dual-geometry}. \Cref{tab:dual-geometry} contrasts, for each of four models, the mean Gini of 1{,}000 random token-pair differences against that of the concept-specific pairs: concept pairs concentrate more strongly in high-variance eigendirections than random pairs drawn from the same vocabulary, in every case at $p<10^{-4}$. Across the 11 models for which single-token unembedding pairs are available (\Cref{fig:method-comparison}) the mean unembedding Gini is $+0.090$; there $N=296$ rather than $297$ because SmolLM2-1.7B contributes 26 concepts, one pair having no single-token form.

\begin{table}[t]
\centering
\caption{Dual geometry: concept-specific unembedding-row pairs concentrate \emph{more than random} token pairs.}
\label{tab:dual-geometry}
\small
\resizebox{\columnwidth}{!}{%
\begin{tabular}{lccc}
\toprule
\textbf{Model} & \textbf{Random Gini} & \textbf{Concept Gini} & \textbf{$p$} \\
\midrule
Qwen2.5-0.5B   & $-0.095$ & $+0.046$ & $< 10^{-4}$ \\
Llama-3.2-1B   & $-0.124$ & $+0.103$ & $< 10^{-4}$ \\
Gemma-2-2b     & $-0.117$ & $+0.073$ & $< 10^{-4}$ \\
SmolLM2-1.7B   & $-0.071$ & $+0.056$ & $< 10^{-4}$ \\
\bottomrule
\end{tabular}%
}
\end{table}

\section{Scaling Data}
\label{app:scaling}

\begin{table*}[t]
\centering
\caption{Qwen~2.5 and Qwen~3 scaling. Gap is concept \scm minus random; Gini gap is concept minus random.}
\label{tab:scaling-full}
\small
\begin{tabular}{lcccccc}
\toprule
\textbf{Model} & \textbf{Concept \scm} & \textbf{Random \scm} & \textbf{Gap} & \textbf{Concept Gini} & \textbf{Random Gini} & \textbf{Gini Gap} \\
\midrule
\multicolumn{7}{l}{\textit{Qwen 2.5}} \\
0.5B  & 0.997 & 0.711 & $+$0.286 & $-$0.356 & $-$0.161 & $-$0.195 \\
1.5B  & 0.986 & 0.729 & $+$0.257 & $-$0.282 & $-$0.172 & $-$0.110 \\
3B    & 0.973 & 0.735 & $+$0.238 & $-$0.272 & $-$0.175 & $-$0.097 \\
7B    & 0.999 & 0.757 & $+$0.242 & $-$0.396 & $-$0.186 & $-$0.210 \\
14B   & 0.996 & 0.790 & $+$0.205 & $-$0.346 & $-$0.207 & $-$0.139 \\
\midrule
\multicolumn{7}{l}{\textit{Qwen 3}} \\
0.6B  & 0.938 & 0.723 & $+$0.215 & $-$0.216 & $-$0.170 & $-$0.046 \\
1.7B  & 0.997 & 0.738 & $+$0.259 & $-$0.248 & $-$0.177 & $-$0.070 \\
4B    & 0.994 & 0.750 & $+$0.243 & $-$0.312 & $-$0.186 & $-$0.126 \\
8B    & 1.000 & 0.783 & $+$0.217 & $-$0.359 & $-$0.207 & $-$0.152 \\
\bottomrule
\end{tabular}
\end{table*}

\begin{figure*}[t]
  \centering
  \begin{subfigure}[b]{0.49\textwidth}
    \includegraphics[width=\textwidth]{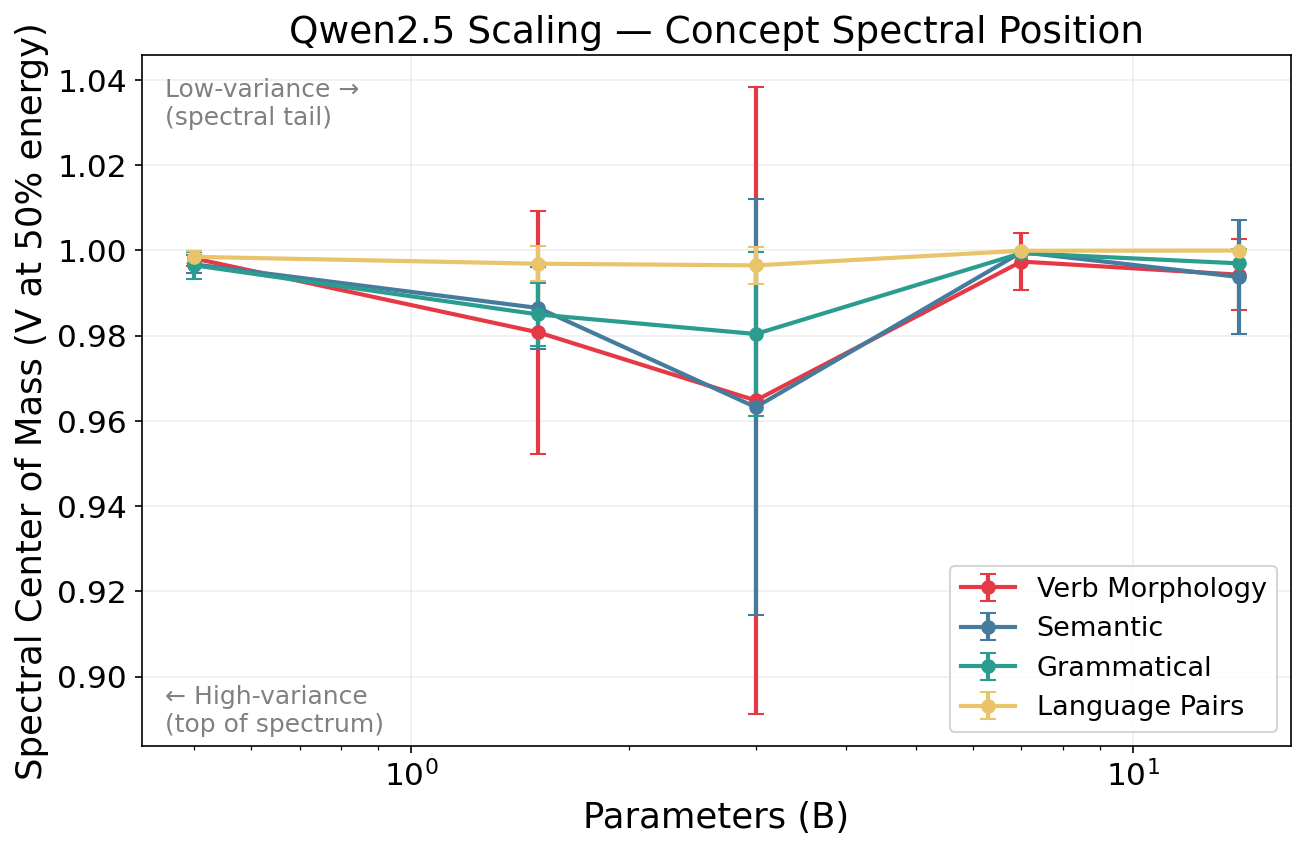}
    \caption{Qwen~2.5 (0.5B--14B)}
  \end{subfigure}\hfill
  \begin{subfigure}[b]{0.49\textwidth}
    \includegraphics[width=\textwidth]{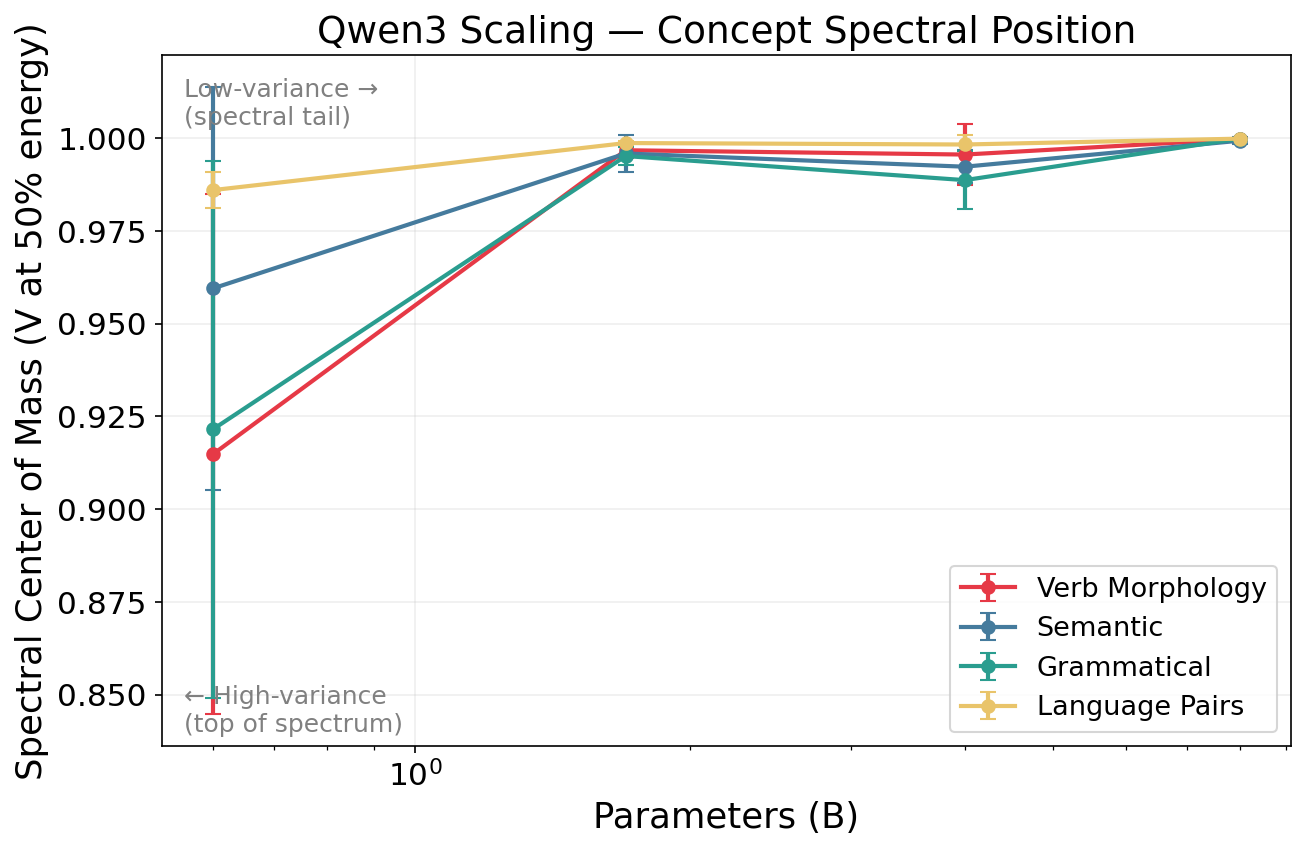}
    \caption{Qwen~3 (0.6B--8B)}
  \end{subfigure}
  \caption{Scaling behavior of spectral anti-concentration.}
  \label{fig:scaling}
\end{figure*}

\begin{figure*}[t]
  \centering
  \begin{subfigure}[b]{\textwidth}
    \centering
    \includegraphics[width=\textwidth]{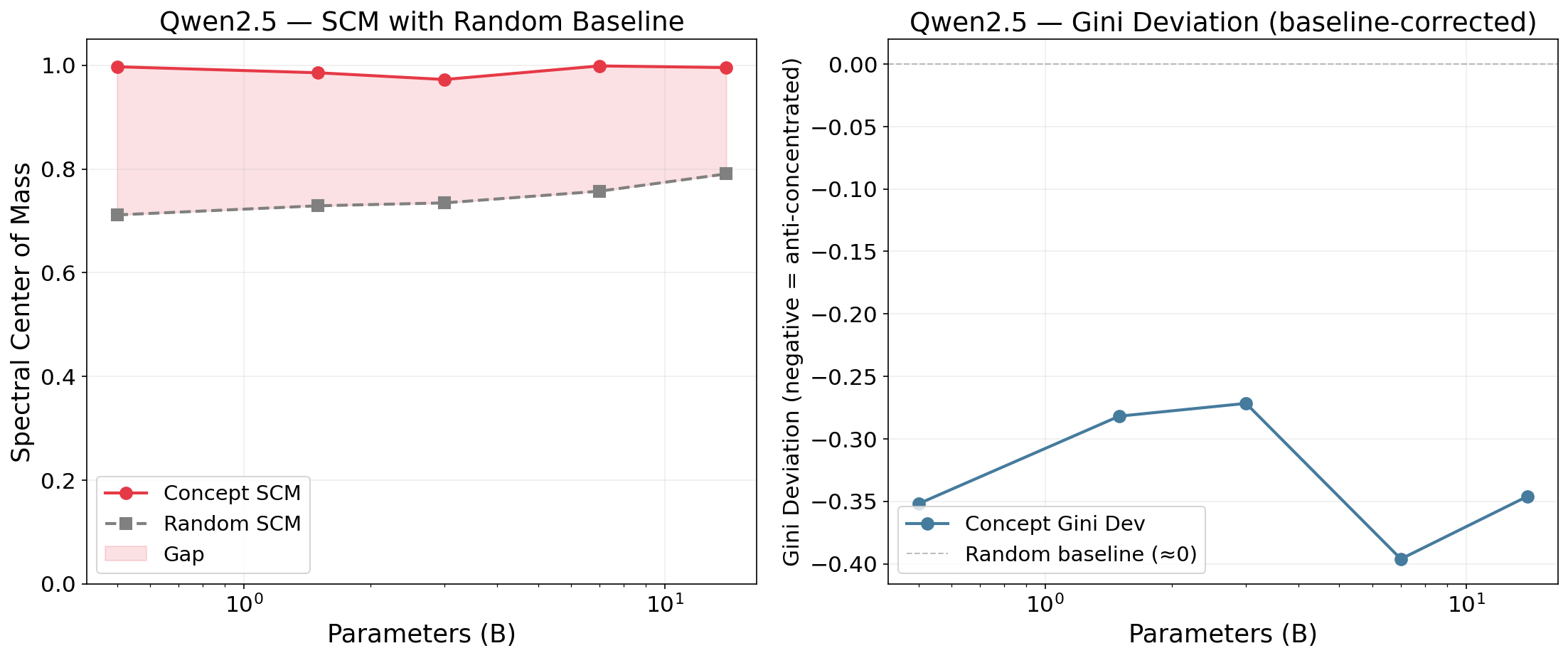}
    \caption{Qwen~2.5 (baseline-corrected)}
  \end{subfigure}
  \par\medskip
  \begin{subfigure}[b]{\textwidth}
    \centering
    \includegraphics[width=\textwidth]{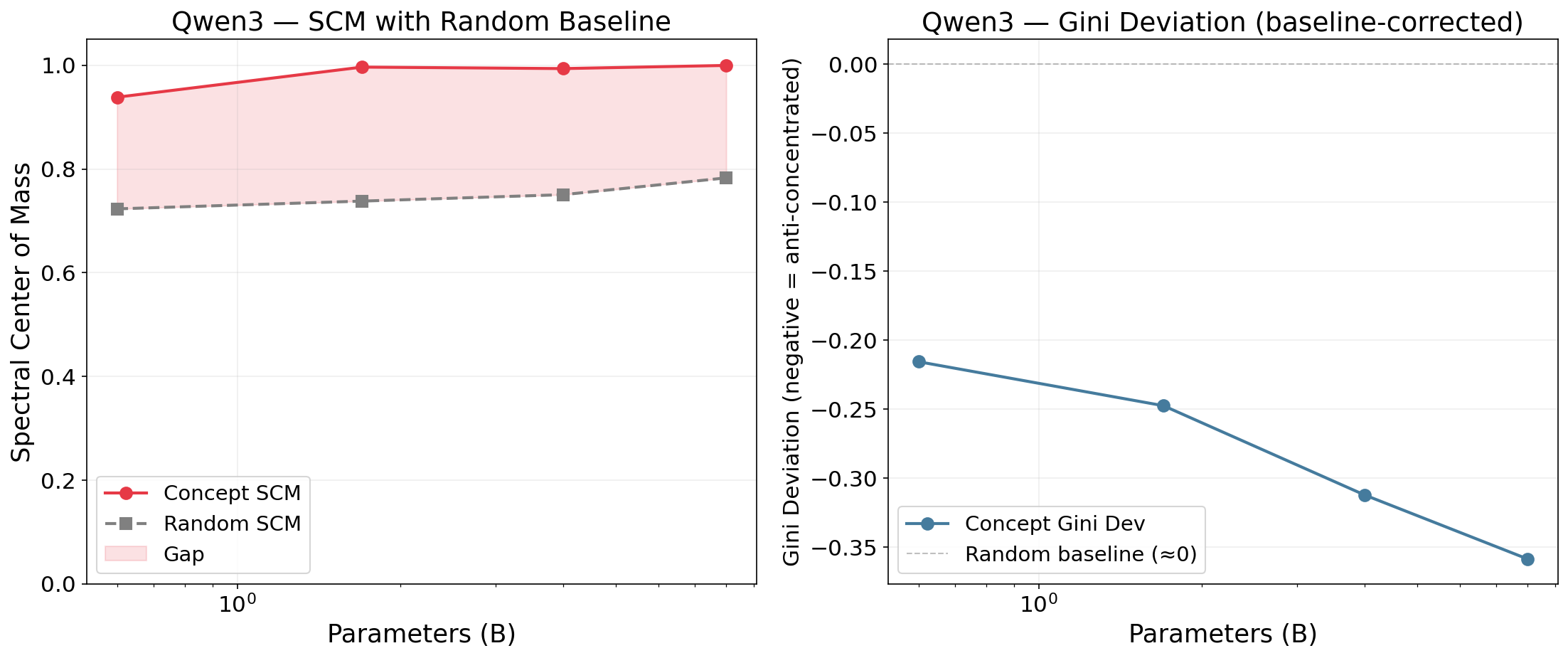}
    \caption{Qwen~3 (baseline-corrected)}
  \end{subfigure}
  \caption{Baseline-corrected scaling. Concept (red) vs.\ random (grey) \scm gaps persist across scales; Gini deviations are stable rather than emerging or vanishing with capacity.}
  \label{fig:scaling-corrected}
\end{figure*}

\Cref{tab:scaling-full} lists the per-scale figures and \Cref{fig:scaling,fig:scaling-corrected} plot them, raw and baseline-corrected. The \scm gap is non-monotonic in capacity; the Gini gap widens monotonically with size in Qwen~3 ($-0.046$ to $-0.152$) but not in Qwen~2.5, where the 0.5B and 7B models show the widest gaps. We interpret this as evidence that anti-concentration is present at all tested scales rather than emerging or disappearing with capacity.

\section{Architecture Comparison}
\label{app:architecture}

\begin{figure*}[t]
  \centering
  \includegraphics[width=\textwidth]{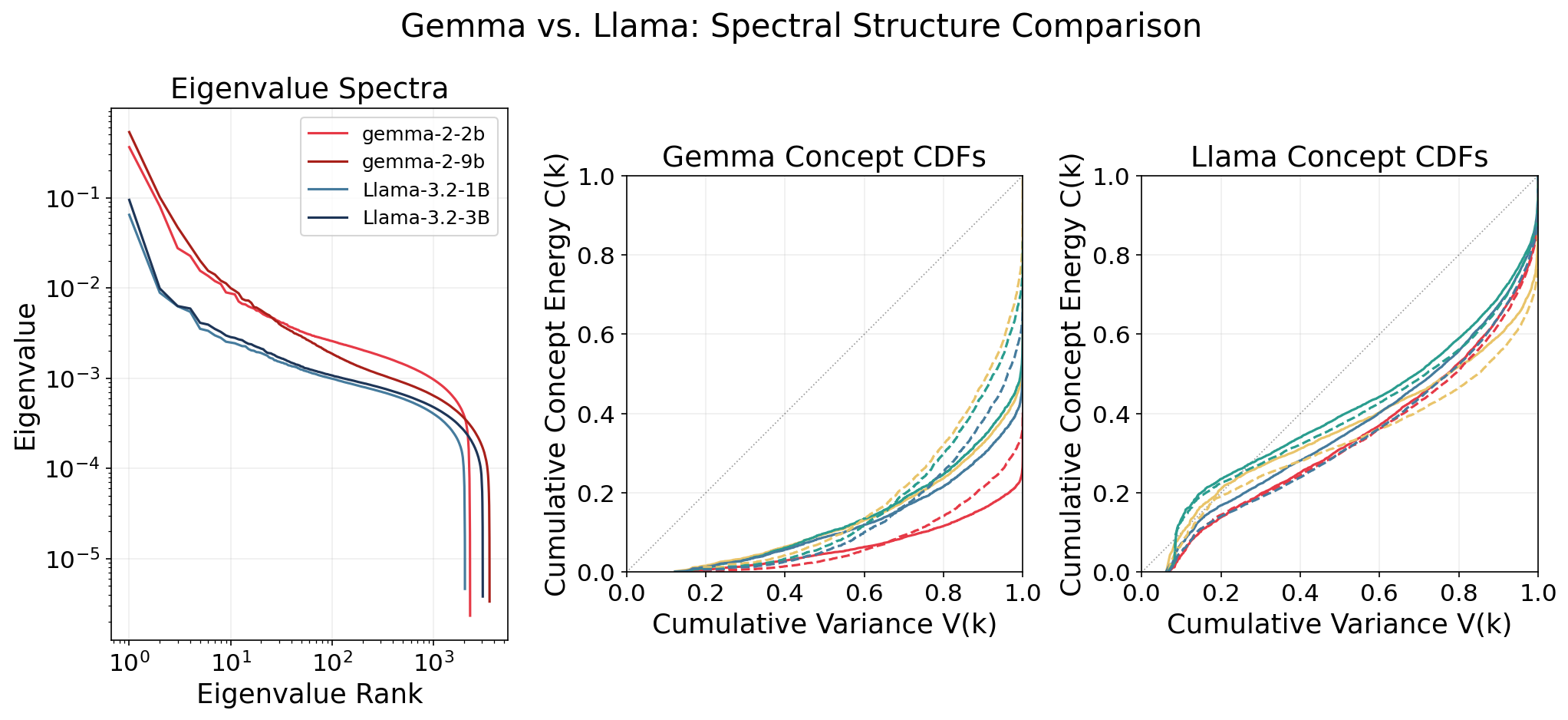}
  \caption{Architecture comparison: Gemma vs.\ Llama spectral energy CDFs. Despite Gemma's more aggressive eigenvalue concentration, both families show comparable anti-concentration of concept vectors.}
  \label{fig:gemma-vs-llama}
\end{figure*}

Gemma models exhibit the highest eigenvalue concentration (top 10\% captures roughly 40--52\% of variance for gemma-2-2b and gemma-2-9b, vs.\ 30--32\% for Llama-3.2-1B and Llama-3.2-3B), consistent with their larger vocabulary (256K vs.\ 128K). Despite this difference, both families show strong anti-concentration; MoE models (OLMoE, JetMoE) exhibit similar patterns, suggesting the phenomenon is not specific to dense FFNs versus sparse experts. \Cref{fig:gemma-vs-llama} overlays the two families' spectral energy CDFs.

\section{Split-Injection Details}
\label{app:hole4-details}

\begin{figure*}[t]
  \centering
  \includegraphics[width=\textwidth]{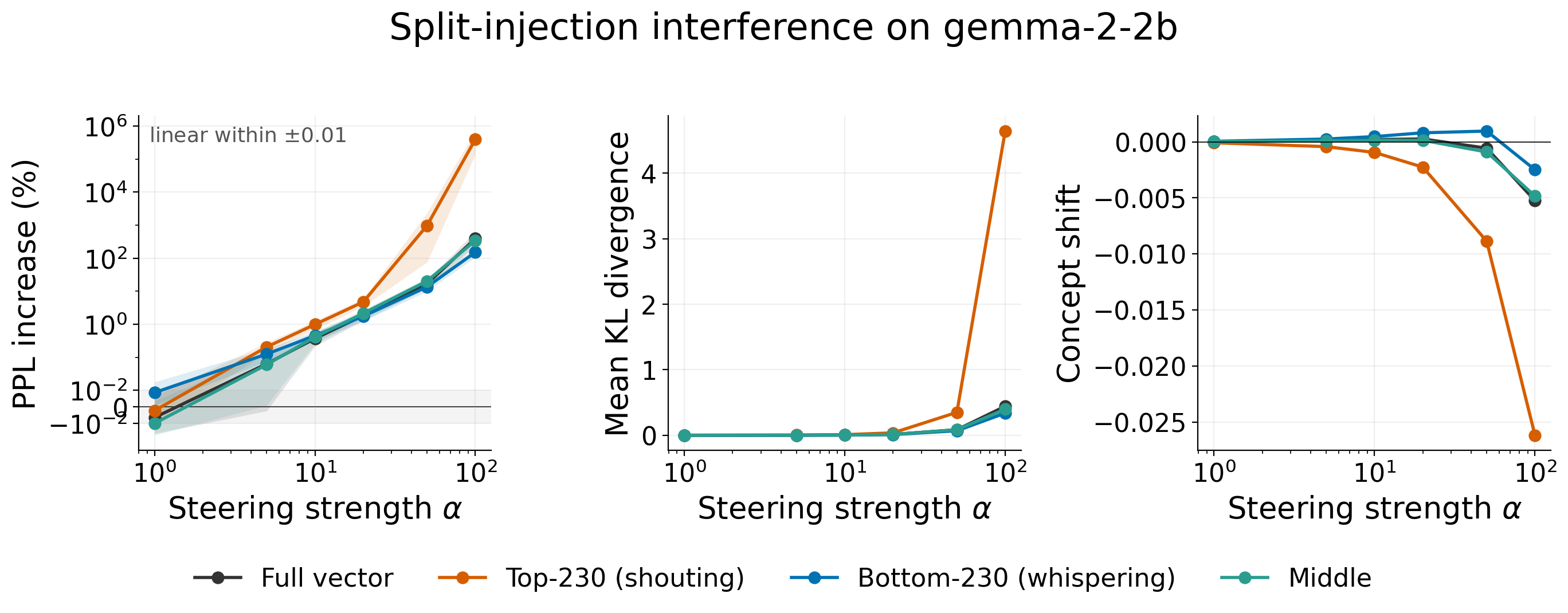}
  \caption{Split-injection on gemma-2-2b: PPL increase, KL divergence, and concept shift vs.\ $\alpha$. At $\alpha=10$ and $\alpha=20$---both inside the validity window---the shouting component causes more PPL degradation than the whispering component; at $\alpha=1$ the difference is not significant and runs the other way ($d_z=-0.29$), and at $\alpha=100$ the shouting arm leaves the validity window ($+401$K\%).}
  \label{fig:split-injection-gemma}
\end{figure*}

\begin{figure*}[t]
  \centering
  \begin{subfigure}[b]{\textwidth}
    \centering
    \includegraphics[width=\textwidth]{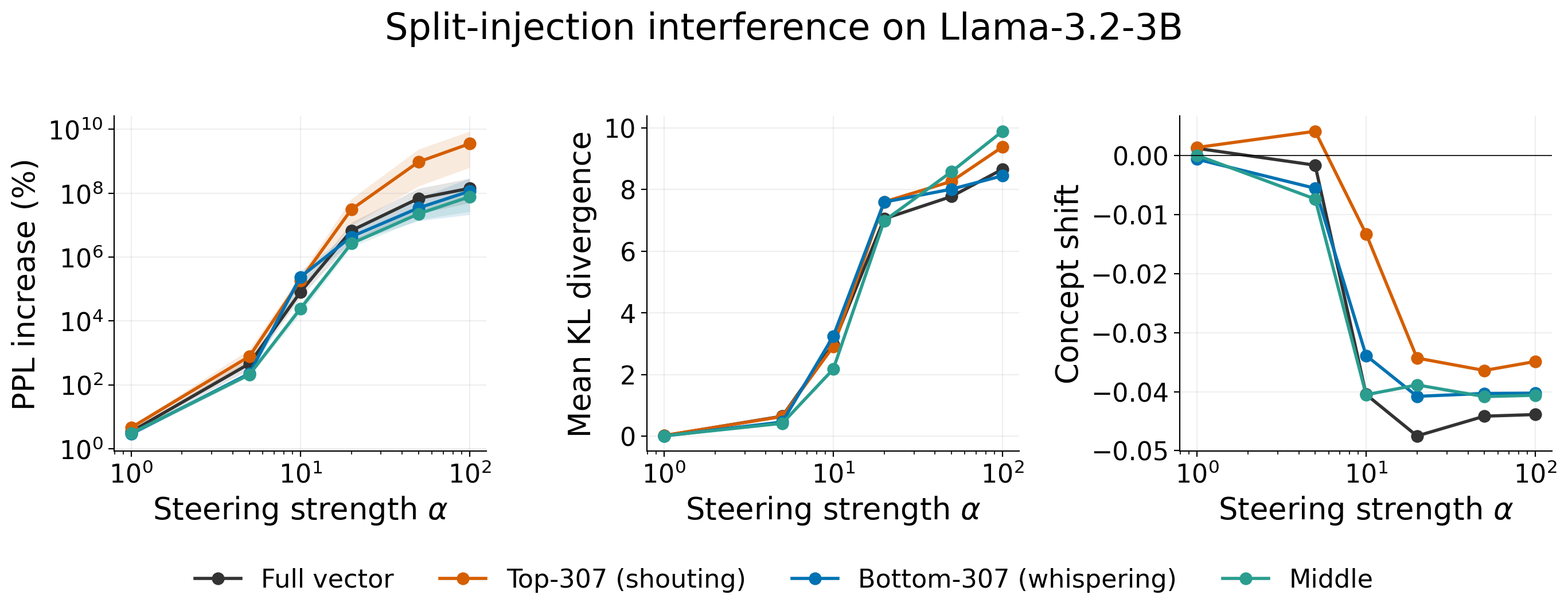}
    \caption{Llama-3.2-3B}
  \end{subfigure}
  \par\medskip
  \begin{subfigure}[b]{\textwidth}
    \centering
    \includegraphics[width=\textwidth]{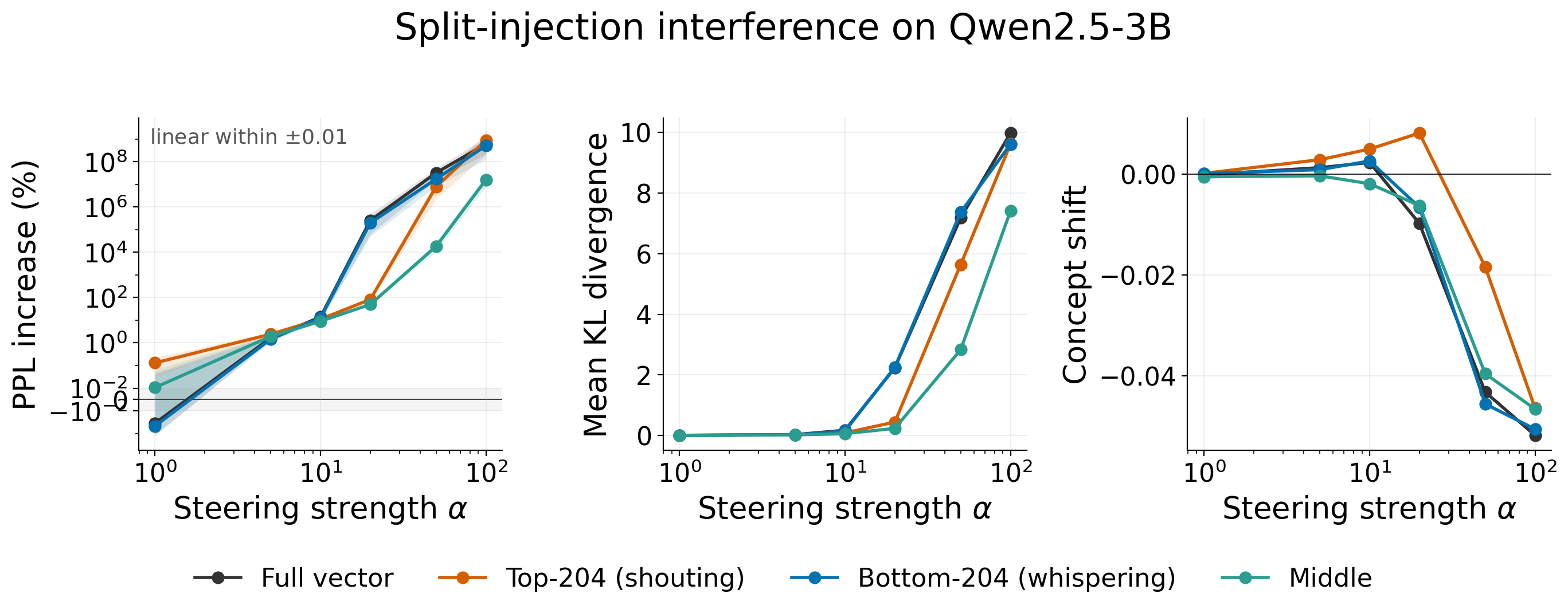}
    \caption{Qwen2.5-3B}
  \end{subfigure}
  \par\medskip
  \begin{subfigure}[b]{\textwidth}
    \centering
    \includegraphics[width=\textwidth]{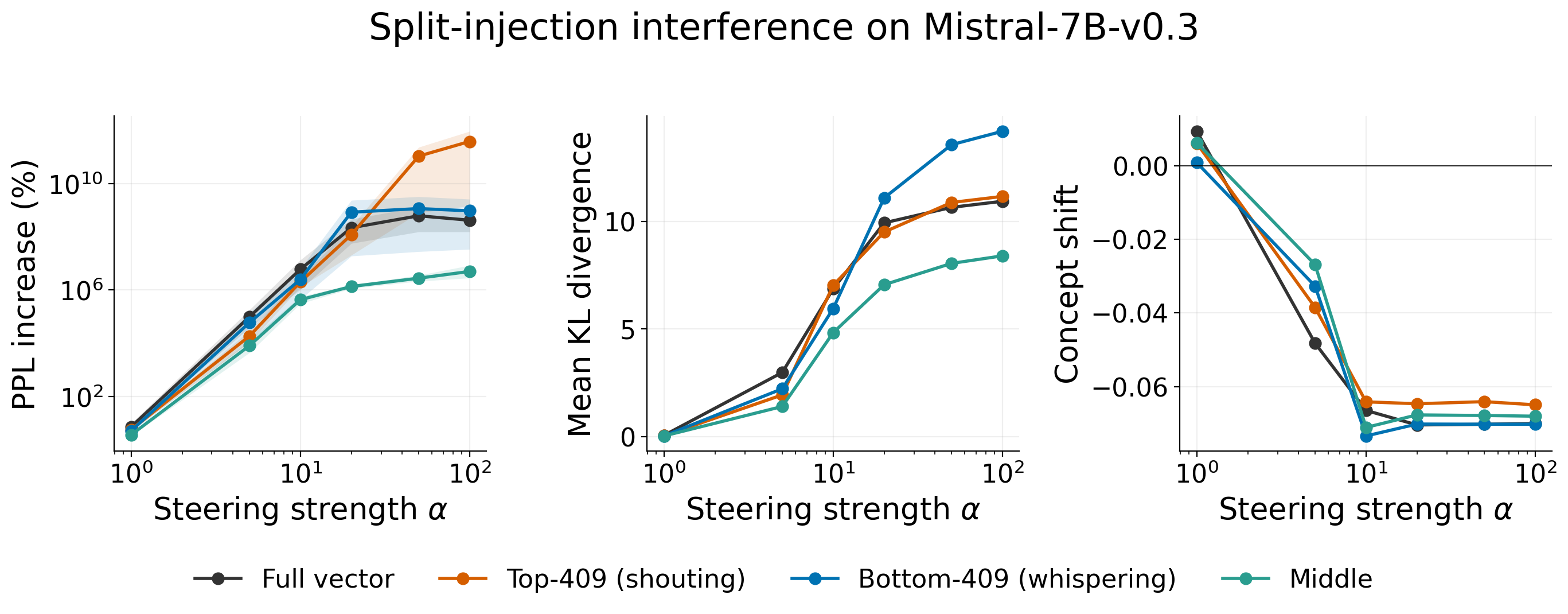}
    \caption{Mistral-7B-v0.3}
  \end{subfigure}
  \caption{Remaining split-injection panels. Llama-3.2-3B and Qwen2.5-3B both show the shouting-greater-than-whispering asymmetry at low $\alpha$, inside the validity window. On Qwen2.5-3B whispering appears to cause \emph{more} disruption at $\alpha=20$ and $50$, but at least one arm has collapsed there, placing those configurations outside the window; Mistral-7B has a single in-window configuration and shows no significant asymmetry (\Cref{tab:split-injection-full}).}
  \label{fig:split-injection-appendix}
\end{figure*}

\Cref{fig:split-injection-gemma} shows the gemma-2-2b panels and \Cref{fig:split-injection-appendix} the remaining three models. All five split-injection runs use the full concept set: every model, gemma-2-2b included, contributes $N=27$ concepts at every $\alpha$. Separately, several concepts on gemma-2-2b have multi-token word pairs producing zero-norm activation-derived vectors and zero spectral energy; those concepts are excluded from the purely spectral analyses on gemma---for instance the norm-matched-null experiment, where gemma-2-2b's effective $N$ is 18---but not from split-injection, which measures perplexity under steering.

\begin{table*}[t]
\centering
\caption{\textbf{Complete split-injection $\alpha$ sweep}: all six $\alpha$ values for all five models (30 configurations); nothing is omitted. Shout/Whisper give the mean perplexity increase under injection along the top-$k$ and bottom-$k$ components. $d_z$ is the paired Cohen's $d$ over the 27 per-concept shout$-$whisper differences, matching the paired $t$-test $p$. A \checkmark marks the 13 configurations inside the validity window (both arms below a $+1000\%$ PPL increase; \S\ref{sec:why}); rows without it are uninterpretable because at least one arm has collapsed, and are shown for completeness only. Every significant in-window row is positive (shouting disrupts more), and the sweep's single significant reversal (Qwen2.5-3B at $\alpha=20$) falls outside the window. Within the window the magnitude still varies with $\alpha$, so no single $\alpha$ characterizes a model.}
\label{tab:split-injection-full}
\footnotesize
\begin{tabular}{llccccc}
\toprule
\textbf{Model} & \textbf{$\alpha$} & \textbf{Shout PPL} & \textbf{Whisper PPL} & \textbf{$d_z$} & \textbf{$p$} & \textbf{In win.} \\
\midrule
\multirow{6}{*}{gemma-2-2b} & 1 & $-0.0\%$ & $+0.0\%$ & $-0.29$ & .1459 & \checkmark \\
 & 5 & $+0.2\%$ & $+0.1\%$ & $+0.37$ & .0674 & \checkmark \\
 & 10 & $+1.0\%$ & $+0.5\%$ & $+0.85$ & .0002 & \checkmark \\
 & 20 & $+4.8\%$ & $+1.8\%$ & $+1.19$ & $<\!10^{-4}$ & \checkmark \\
 & 50 & $+982\%$ & $+14\%$ & $+0.30$ & .1260 & \checkmark \\
 & 100 & $+401$K$\%$ & $+153\%$ & $+0.52$ & .0125 &  \\
\midrule
\multirow{6}{*}{Llama-3.1-8B} & 1 & $+5.6\%$ & $+4.7\%$ & $+0.38$ & .0573 & \checkmark \\
 & 5 & $+383\%$ & $+223\%$ & $+1.09$ & $<\!10^{-4}$ & \checkmark \\
 & 10 & $+230$K$\%$ & $+90$K$\%$ & $+0.26$ & .1826 &  \\
 & 20 & $+238$M$\%$ & $+16$M$\%$ & $+0.68$ & .0015 &  \\
 & 50 & $+5.7$B$\%$ & $+39$M$\%$ & $+0.71$ & .0011 &  \\
 & 100 & $+21.8$B$\%$ & $+50$M$\%$ & $+0.56$ & .0073 &  \\
\midrule
\multirow{6}{*}{Llama-3.2-3B} & 1 & $+4.6\%$ & $+2.9\%$ & $+0.90$ & .0001 & \checkmark \\
 & 5 & $+779\%$ & $+225\%$ & $+0.46$ & .0235 & \checkmark \\
 & 10 & $+186$K$\%$ & $+236$K$\%$ & $-0.17$ & .3937 &  \\
 & 20 & $+30.8$M$\%$ & $+4.3$M$\%$ & $+0.31$ & .1139 &  \\
 & 50 & $+956$M$\%$ & $+35$M$\%$ & $+0.26$ & .1828 &  \\
 & 100 & $+3.5$B$\%$ & $+115$M$\%$ & $+0.27$ & .1660 &  \\
\midrule
\multirow{6}{*}{Qwen2.5-3B} & 1 & $+0.1\%$ & $-0.0\%$ & $+0.55$ & .0085 & \checkmark \\
 & 5 & $+2.4\%$ & $+1.5\%$ & $+0.60$ & .0045 & \checkmark \\
 & 10 & $+11\%$ & $+13\%$ & $-0.13$ & .5054 & \checkmark \\
 & 20 & $+82\%$ & $+197$K$\%$ & $-0.43$ & .0350 &  \\
 & 50 & $+7.8$M$\%$ & $+17.2$M$\%$ & $-0.27$ & .1674 &  \\
 & 100 & $+875$M$\%$ & $+503$M$\%$ & $+0.14$ & .4889 &  \\
\midrule
\multirow{6}{*}{Mistral-7B-v0.3} & 1 & $+5.6\%$ & $+5.1\%$ & $+0.31$ & .1168 & \checkmark \\
 & 5 & $+18$K$\%$ & $+59$K$\%$ & $-0.23$ & .2515 &  \\
 & 10 & $+2.0$M$\%$ & $+2.5$M$\%$ & $-0.05$ & .8068 &  \\
 & 20 & $+119$M$\%$ & $+839$M$\%$ & $-0.19$ & .3390 &  \\
 & 50 & $+107.6$B$\%$ & $+1.1$B$\%$ & $+0.34$ & .0915 &  \\
 & 100 & $+384$B$\%$ & $+946$M$\%$ & $+0.30$ & .1371 &  \\
\bottomrule
\end{tabular}
\end{table*}

\section{POS Probing Details}
\label{app:pos-probe}

\begin{table*}[t]
\centering
\caption{Full POS probing results. Variance columns: fraction of total eigenspectrum variance captured by each subspace. Random-$k$ is the mean accuracy over 5 random subspaces of the same dimensionality.}
\label{tab:pos-probe-full}
\footnotesize
\begin{tabular}{lccccccccc}
\toprule
\textbf{Model} & \textbf{$d$} & \textbf{$k$} & \textbf{Top-$k$ var.} & \textbf{Bot-$k$ var.} & \textbf{Top-$k$} & \textbf{Bot-$k$} & \textbf{Rand-$k$} & \textbf{Full} & \textbf{Gap} \\
\midrule
gemma-2-2b     & 2304 & 230 & 39.8\% & 1.6\% & .554 & .506 & .526 & .813 & $+.048$ \\
Llama-3.2-3B   & 3072 & 307 & 32.0\% & 2.1\% & .628 & .576 & .584 & .835 & $+.052$ \\
Llama-3.1-8B   & 4096 & 409 & 27.5\% & 1.8\% & .641 & .625 & .602 & .833 & $+.016$ \\
Mistral-7B     & 4096 & 409 & 31.3\% & 1.5\% & .624 & .607 & .606 & .859 & $+.017$ \\
SmolLM2-1.7B   & 2048 & 204 & 27.4\% & 2.3\% & .560 & .536 & .537 & .781 & $+.024$ \\
Qwen3-4B       & 2560 & 256 & 28.4\% & 2.4\% & .572 & .555 & .562 & .826 & $+.018$ \\
\midrule
Qwen2.5-3B     & 2048 & 204 & 27.0\% & 2.6\% & .557 & .576 & .555 & .787 & $-.019$ \\
Qwen2.5-7B     & 3584 & 358 & 26.2\% & 1.9\% & .572 & .586 & .575 & .813 & $-.014$ \\
\bottomrule
\end{tabular}
\end{table*}

\begin{table*}[t]
\centering
\caption{Chinese UD probing on Qwen models. Both show bottom-$k$ $>$ top-$k$, so the Qwen~2.5 English reversal is not explained solely by applying an English-derived covariance to a multilingual model.}
\label{tab:pos-probe-chinese}
\small
\begin{tabular}{lccccccc}
\toprule
\textbf{Model} & \textbf{$k$} & \textbf{Top-$k$} & \textbf{Bot-$k$} & \textbf{Rand-$k$} & \textbf{Full} & \textbf{Gap} & \textbf{$p$} \\
\midrule
Qwen2.5-3B & 204 & .411 & .462 & .427 & .692 & $-.051$ & .0001 \\
Qwen3-4B   & 256 & .432 & .446 & .449 & .707 & $-.013$ & .014 \\
\bottomrule
\end{tabular}
\end{table*}

\begin{figure*}[t]
  \centering
  \includegraphics[width=\textwidth]{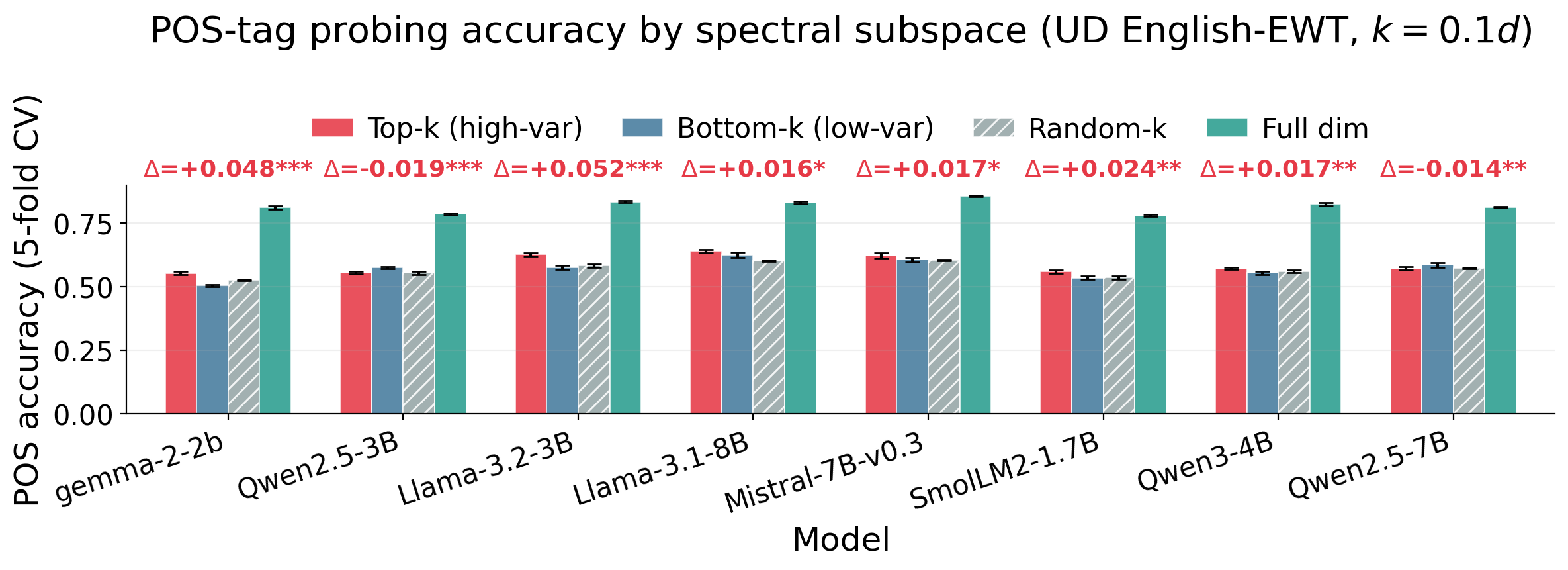}
  \caption{POS-tag probing across 8 models. Top-$k$ (red) consistently outperforms bottom-$k$ (blue) except on Qwen~2.5. Full-dimensional probes (green) serve as upper bounds (78--86\%).}
  \label{fig:pos-probe}
\end{figure*}

\begin{figure*}[t]
  \centering
  \includegraphics[width=0.9\textwidth]{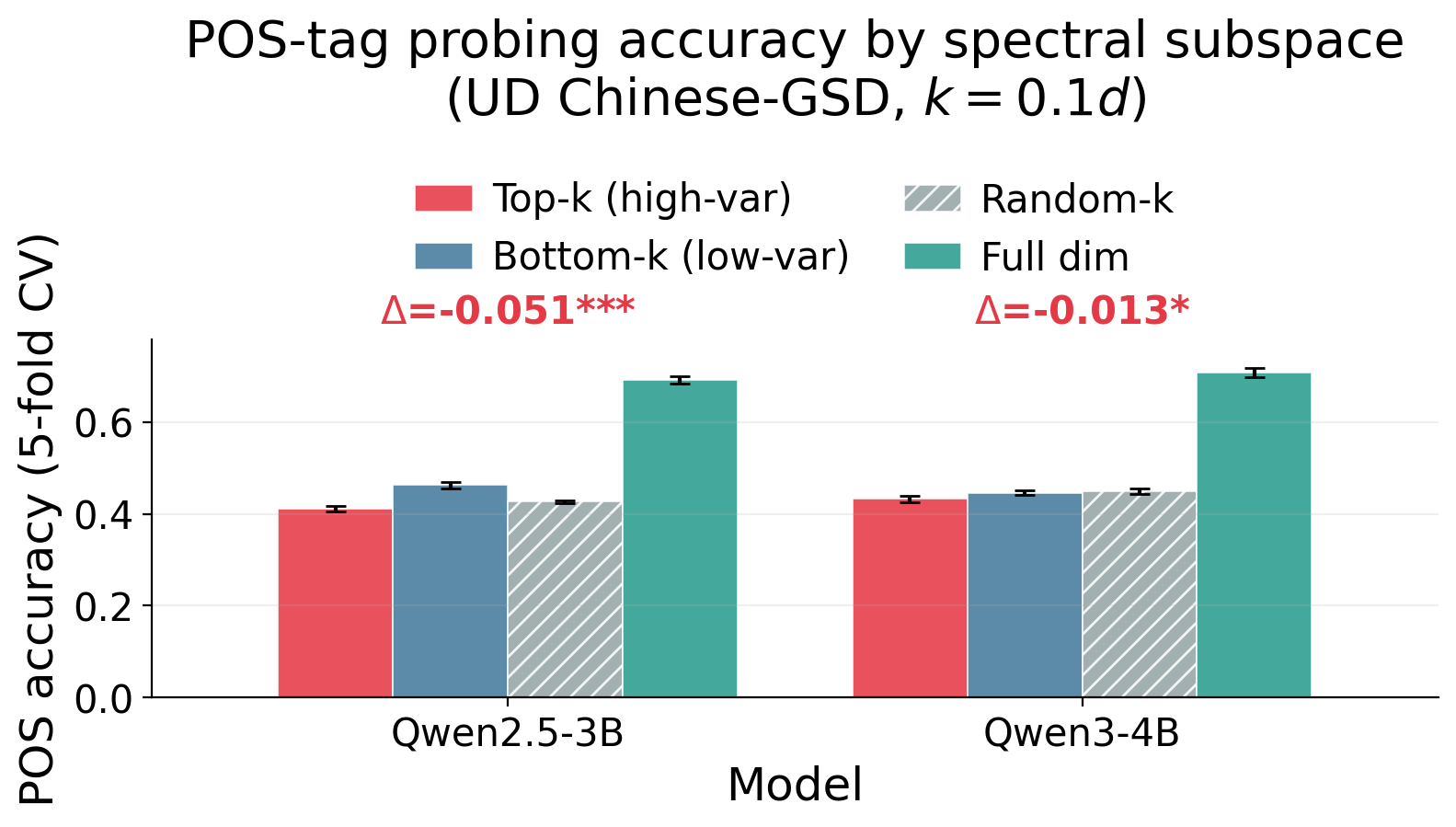}
  \caption{Chinese POS probing on Qwen models: both show bottom-$k$ $>$ top-$k$.}
  \label{fig:pos-probe-chinese}
\end{figure*}

\Cref{tab:pos-probe-full} gives the full English results and \Cref{tab:pos-probe-chinese} the Chinese control; \Cref{fig:pos-probe,fig:pos-probe-chinese} plot them. The top-$k$ subspace captures 26--40\% of total variance (vs.\ $k/d=10\%$), while bottom-$k$ captures only 1.5--2.6\%. Despite this asymmetry the accuracy gaps are modest (1.6--5.2 pp on positive-gap models), indicating syntactic information is distributed but biased toward the high-variance subspace.

\section{POS Probing Sensitivity Analysis}
\label{app:pos-sensitivity}

To confirm that the POS-probing result (\Cref{sec:why}) is not an artifact of the specific spectral cutoff, we repeat the top-$k$ vs.\ bottom-$k$ analysis at $k\in\{5\%,10\%,20\%\}$ for all eight models. \Cref{tab:pos-sensitivity} reports the top-$k$ and bottom-$k$ accuracies, the top-minus-bottom gap, and the paired $t$-test $p$-value at each cutoff. The \emph{direction} of the gap is invariant to the cutoff: the six models with a positive (high-variance) gap retain it at every $k$, and the Qwen~2.5 reversal persists at all three cutoffs---including $k=20\%$---ruling out the possibility that a single cutoff is simply too coarse to separate the two subspaces.

\begin{table*}[t]
\centering
\caption{POS-tag probing across spectral cutoffs $k\in\{5\%,10\%,20\%\}$: Top-$k$ and Bottom-$k$ accuracies, Gap, and paired $t$-test $p$. The qualitative direction of the gap is invariant across cutoffs for all eight models.}
\label{tab:pos-sensitivity}
\resizebox{\textwidth}{!}{
\begin{tabular}{lcccc|cccc|cccc}
\toprule
& \multicolumn{4}{c|}{\textbf{$k=5\%$}} & \multicolumn{4}{c|}{\textbf{$k=10\%$}} & \multicolumn{4}{c}{\textbf{$k=20\%$}} \\
\textbf{Model} & \textbf{Top} & \textbf{Bot} & \textbf{Gap} & \textbf{$p$} & \textbf{Top} & \textbf{Bot} & \textbf{Gap} & \textbf{$p$} & \textbf{Top} & \textbf{Bot} & \textbf{Gap} & \textbf{$p$} \\
\midrule
gemma-2-2b & .494 & .442 & $+.051$ & .0005 & .554 & .506 & $+.048$ & .0001 & .624 & .592 & $+.031$ & .0006 \\
Llama-3.2-3B & .565 & .501 & $+.064$ & .0008 & .628 & .576 & $+.052$ & .0001 & .688 & .659 & $+.028$ & .0075 \\
Llama-3.1-8B & .579 & .566 & $+.013$ & .0634 & .641 & .625 & $+.016$ & .0121 & .703 & .690 & $+.013$ & .0653 \\
Mistral-7B-v0.3 & .566 & .554 & $+.013$ & .0506 & .624 & .607 & $+.017$ & .0125 & .681 & .653 & $+.028$ & .0009 \\
SmolLM2-1.7B & .493 & .482 & $+.011$ & .0503 & .560 & .536 & $+.024$ & .0023 & .604 & .582 & $+.023$ & .0001 \\
Qwen3-4B & .518 & .495 & $+.023$ & .0052 & .572 & .555 & $+.018$ & .0012 & .629 & .619 & $+.010$ & .0482 \\
\midrule
Qwen2.5-3B & .506 & .517 & $-.011$ & .0344 & .557 & .576 & $-.019$ & $<.0001$ & .611 & .631 & $-.021$ & .0004 \\
Qwen2.5-7B & .516 & .525 & $-.009$ & .1515 & .572 & .586 & $-.014$ & .0015 & .623 & .638 & $-.014$ & .0129 \\
\bottomrule
\end{tabular}
}
\end{table*}

\end{document}